\PassOptionsToPackage{table}{xcolor}
\documentclass{article}
\usepackage{iclr2027_conference}
\usepackage{times}

\usepackage{amsmath,amsfonts,bm}

\def\eqref#1{equation~\ref{#1}}

\def\1{\bm{1}}

\DeclareMathAlphabet{\mathsfit}{\encodingdefault}{\sfdefault}{m}{sl}
\SetMathAlphabet{\mathsfit}{bold}{\encodingdefault}{\sfdefault}{bx}{n}

\newcommand{\R}{\mathbb{R}}

\newcommand{\sigmoid}{\sigma}

\usepackage{amsmath}
\usepackage{amssymb}
\usepackage{amsthm}
\usepackage{booktabs}
\usepackage{graphicx}
\usepackage{microtype}
\usepackage{placeins}
\usepackage{tabularx}
\usepackage{xspace}
\usepackage{url}
\usepackage{hyperref}
\hypersetup{hidelinks}


\title{GRAFT-EEG: Gauge-Aware Physical Field Transport\\
for Cross-Dataset Motor-Imagery EEG}

\newcommand{\method}{\textsc{Graft-EEG}\xspace}

\newcommand{\Omat}{\mathcal{O}}
\newcommand{\hashsplit}[8]{\texttt{\footnotesize #1\allowbreak#2\allowbreak#3\allowbreak#4\allowbreak#5\allowbreak#6\allowbreak#7\allowbreak#8}}
\definecolor{GraftBlue}{HTML}{0072B2}
\definecolor{GraftTeal}{HTML}{009E73}
\definecolor{GraftOrange}{HTML}{D55E00}
\definecolor{GraftPurple}{HTML}{6B5AA6}
\definecolor{GraftGray}{HTML}{6C757D}
\newcommand{\partialden}[1]{\cellcolor{GraftOrange!12}\textbf{#1}}

\newcommand{\provineligible}[1]{\cellcolor{GraftPurple!10}\textbf{#1}}
\newtheorem{proposition}{Proposition}

\author{
Ying Zhou$^{1}$ \qquad Mengting Liu$^{1}$\thanks{Corresponding author.}\\
$^{1}$School of Biomedical Engineering, Shenzhen Campus of Sun Yat-sen University
}

\iclrfinalcopy

\begin{document}
\maketitle

\fancyhead{}
\renewcommand{\headrulewidth}{0pt}

\begin{abstract}
Cross-dataset EEG decoding often treats montage, reference, and active-sensor changes as representation shift, although they also alter the observation operator. We study target-statistics-blind transfer: fitting receives no target signal, label, or empirical statistic, and electrode coordinates only instantiate a fixed prediction operator. \method removes the active-set common-reference gauge and transports each native mask through a template motor leadfield. A source-fitted spectral classifier anchors the decision; a bounded state-space-model (SSM) residual is secondary. We establish restricted identifiability and an operator-perturbation bound. Across five leave-one-dataset-out targets (253 participants), dynamic \method achieved 60.865\% equal-target balanced accuracy versus 52.684\% for strict-clean 5/5 BIOT (hierarchical difference $+8.182$ points; 95\% CI $+6.382$ to $+10.196$; positive in every target). On four EEGPT-clean targets, the descriptive means were 61.520\% and 59.057\%. In four post-hoc exact same-parent tests, FULL exceeded coordinate-basis, matched-spectrum, row-shuffled, and wrong-anatomy controls by 9.959, 9.698, 11.241, and 3.985 points, respectively (all 95\% CIs excluded zero; Holm-adjusted $P<0.001$). A separate 60-participant same-paradigm evaluation retained a positive physical-operator contrast ($+4.029$ points; $+2.338$ to $+5.773$), whereas dynamic--static was inconclusive; its preregistered status is partial confirmation rather than unrestricted external validation. The SSM added $+1.059$ points ($+0.013$ to $+1.892$). Claims remain bounded to five binary motor-imagery domains and the tested operators.
\end{abstract}

\section{Introduction}

Public motor-imagery EEG datasets differ in electrode count, location, reference, amplifier, sessions, and unusable channels. Transfer therefore couples physiological and procedural shift with a change in observation coordinates. Channel intersection resolves tensor shape by discarding native sensors; learned adapters can retain them but may obscure which target information fitted the adapter.

Large EEG models learn reusable representations through masking, tokenization, and diverse corpora \citep{jiang2024labram,wang2024eegpt,wang2025cbramod,jiang2025neurolm}. BIOT tokenizes channels separately; newer systems use coordinates or learned channel-unification modules \citep{yang2023biot,elouahidi2025reve,doner2025luna,chen2026unintfm}. These advances do not isolate whether a full unseen montage can be decoded before opening any target signal or empirical statistic. We test explicit observation-operator transport under that fixed contract, a distinction that matters when compact BCI decoders remain competitive and conclusions are protocol-sensitive \citep{yang2026worthit}.

In our \emph{target-statistics-blind heterogeneous-montage} contract, declared coordinates may instantiate geometry-only operators, but fitting uses source signals and labels only. Target geometry supplies no covariance, normalization, calibration, validation loss, or stopping signal; after source fitting, target signals are mapped once and labels reach only the scorer. Exact same-parent controls fix split, seed, head, features, ridge rule, and budget while replacing the motor leadfield with coordinate-only, spectrum-preserving random, row-shuffled, or wrong-anatomy operators. A stricter sensitivity learns the shared basis and hyperparameters from source geometries, opening the target montage only after all 25 fits are frozen.

\method first removes the unidentifiable common-reference component on each trial's active channels and transports the remaining field through a mask-specific inverse into a fixed bilateral motor basis. It then separates a static spectral anchor from a small bounded state-space residual, whose zero initialization recovers the static classifier and whose stable decay bounds the recurrent correction.

Our contributions are:
\begin{itemize}
    \item a precise access contract for cross-dataset EEG decoding that distinguishes target geometry from target signal, label, and distributional statistics;
    \item an active-set reference quotient and physical motor-field transport into an eight-dimensional latent, with restricted identifiability, an operator-perturbation bound, a static anchor, and a secondary bounded residual;
    \item a five-domain evaluation with strict and partial denominators, domain-aware uncertainty, four exact same-parent operator controls, baseline-health diagnostics, and a same-paradigm partial-confirmation external evaluation.
\end{itemize}

\section{Related Work}

Large EEG models use masked pretraining, channel tokenization, coordinates, or learned queries to accommodate heterogeneous layouts \citep{yang2023biot,jiang2024labram,wang2024eegpt,wang2025cbramod,jiang2025neurolm,elouahidi2025reve,doner2025luna,chen2026unintfm}. Reported downstream protocols are not uniformly target-statistics-blind, so their results do not answer the same access-constrained question. We study a boundary in which the entire target dataset remains unopened during fitting, except for its declared electrode coordinates. \method is therefore neither a foundation model nor a claim of topology-agnostic inference.

Geometry-based FI/SSI and Riemannian alignment provide strong alternatives to common-channel selection \citep{mellot2024physics}. Related cortical work uses participant-specific surface eigenmodes to constrain EEG/MEG source imaging \citep{wang2026geometry}. Our operator instead uses montage coordinates and a shared template motor basis for decoding; it is not participant-specific source localization. The matched FI/SSI controls receive the same native active channels and are denied target recentering, covariance fitting, and calibration. EEGNet and recent state-space decoders motivate the compact heads \citep{lawhern2018eegnet,ma2026codebrain,ryoo2025possm}, but our temporal branch is restricted to a bounded residual and supports no streaming claim.

\section{Problem Setting and Access Contract}

Let $D_m$ have native channels $C_m$, geometry $Q_m$, and trials $X_i^{(m)}\in\R^{|C_m|\times T}$ with active set $S_i^{(m)}\subseteq C_m$ and label $y_i^{(m)}\in\{0,1\}$. In each fold, one complete dataset $D_{m^\star}$ is held out. Its geometry may be opened to instantiate the declared linear operator, but its signal store, labels, trial queue, and empirical statistics cannot be opened during fitting. Source features, hyperparameters, classifiers, state-space weights, and calibration parameters are committed to a hash-bound manifest before a separate process opens target signals. Labels are materialized only after predictions are frozen.

Thus, \emph{target-statistics-blind} means blind to target signals, labels, and empirical statistics during fitting, not blind to electrode coordinates. After lock, a deterministic per-trial mask instantiates only that trial's operator; masks are never aggregated or used for refitting, normalization, calibration, early stopping, or threshold selection. Source-subject validation remains permitted.

\section{GRAFT-EEG}

\begin{figure*}[t]
  \centering
  \includegraphics[width=\textwidth]{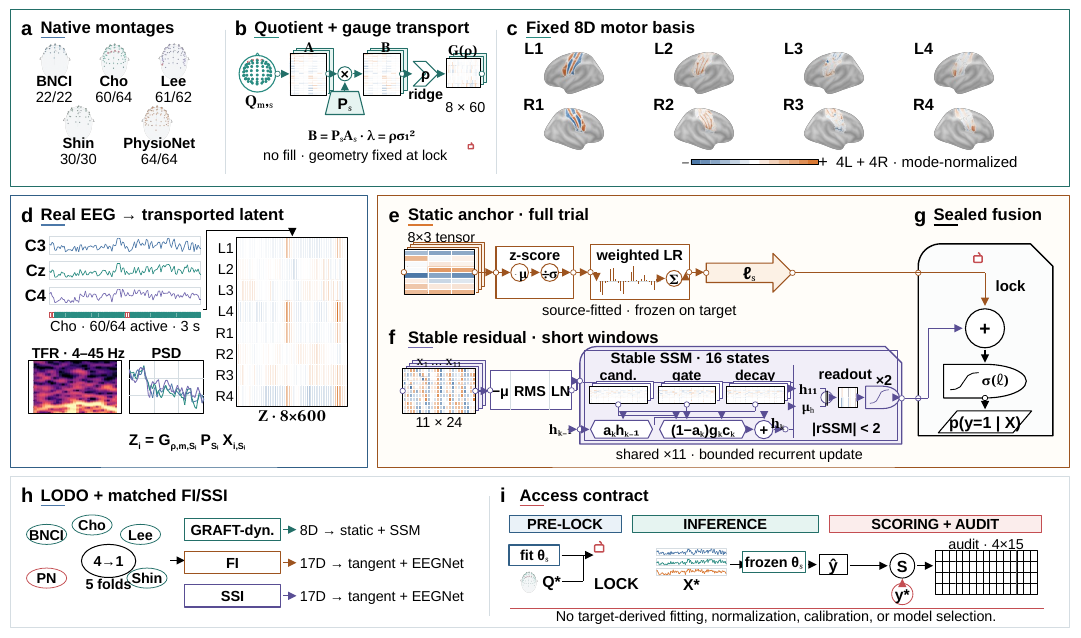}
  \caption{\textbf{GRAFT-EEG and its access contract.} \textbf{a--c}, Quotient reference and transport of native active sensors to a fixed eight-dimensional bilateral motor basis. \textbf{d}, Source-side signals and transported latent. \textbf{e--g}, Static full-trial spectral logit plus a bounded short-window residual before the sigmoid. \textbf{h}, Leave-one-dataset-out evaluation and matched FI/SSI controls. \textbf{i}, Geometry is available before lock, signals only at inference, and labels only at scoring; target-derived fitting, normalization, calibration, and model selection are prohibited.}
  \label{fig:method}
\end{figure*}

Figure~\ref{fig:method} introduces the observation-operator problem and access contract; Figure~\ref{fig:architecture} then separates the geometry-conditioned transport, static anchor, and bounded dynamic residual at the level of the executed model.

\subsection{Active-set reference quotient}

EEG potentials are defined only up to an additive common offset. Applying a fixed inverse directly to differently referenced observations can therefore couple the representation to an arbitrary gauge. For an active set $S$, we use
\begin{equation}
P_S=I_S-\frac{1}{|S|}\mathbf{1}_S\mathbf{1}_S^\top,
\qquad
P_S(x+c\mathbf{1}_S)=P_Sx.
\label{eq:gauge}
\end{equation}
The same $P_S$ is applied to the signal and the matching lead-field rows. Failed channels are excluded rather than filled from target data; Appendix~\ref{app:implementation} gives the fixed retention rules.

\subsection{Mask-specific template motor-field transport}

Let $A_{m,S}\in\R^{|S|\times d}$ be the observation matrix for a fixed $d=8$ template motor basis (four modes per hemisphere). It is constructed once from the declared montage geometries and an fsaverage motor region; no EEG sample, label, or target statistic enters its construction. These coordinates are task-specific features, not participant-specific source estimates. We form
\begin{equation}
B_{m,S}=P_SA_{m,S}, \qquad
\lambda_{m,S}=\rho\,\sigma_1(B_{m,S})^2,
\label{eq:ridge-scale}
\end{equation}
where $\sigma_1$ is the largest singular value and $\rho$ is selected using source-subject validation. The mask-specific inverse and transported latent are
\begin{equation}
G_{m,S}^{\rho}=(B_{m,S}^{\top}B_{m,S}+\lambda_{m,S}I_d)^{-1}B_{m,S}^{\top},
\qquad
Z_i^{(m)}=G_{m,S_i^{(m)}}^{\rho}P_{S_i^{(m)}}X_{i,S_i^{(m)}}^{(m)}.
\label{eq:transport}
\end{equation}
Caching is keyed by dataset, active-mask fingerprint, and $\rho$, so distinct bad-channel patterns cannot share an incompatible inverse. The spectral scale makes $\rho$ comparable across masks.

\subsection{What heterogeneous operators can identify}

The transport above is not a claim of unique nonlinear cortical-source recovery. Its narrower justification can be stated for a finite-dimensional linearized field. Let
\begin{equation}
z_{t+1}=Fz_t, \qquad y_t^{(m)}=A_mz_t+\epsilon_t^{(m)},
\label{eq:linearized-field}
\end{equation}
where $z_t\in\R^d$ is a latent state and $A_m$ is the gauge-projected observation operator for montage $m$. For montages $\mathcal M$ and horizon $T$, define the joint observability matrix
\begin{equation}
\Omat_{\mathcal M,T}=\operatorname{col}\!\left\{A_mF^t:
m\in\mathcal M,\ t=0,\ldots,T-1\right\}.
\label{eq:joint-observability}
\end{equation}

\begin{proposition}[Restricted identifiability]
If $F$ and $\{A_m\}_{m\in\mathcal M}$ are fixed and
$\operatorname{rank}(\Omat_{\mathcal M,T})=d$, noiseless stacked observations uniquely determine $z_0$. If the latent coordinates are learned jointly with the dynamics and operators, the representation is identifiable only up to an invertible similarity transform
$(F,A_m,z_t)\mapsto(SFS^{-1},A_mS^{-1},Sz_t)$.
\label{prop:identifiability}
\end{proposition}

The proposition explains why heterogeneous montages can be complementary: an individual observation operator may have a nontrivial null space, while their jointly observed modes can span the restricted latent. Gauge removal is necessary before applying this argument because a common reference offset is not an identifiable field component. It does not, however, make arbitrary nonlinear sources identifiable.

\begin{proposition}[Perturbed-operator reconstruction]
Let $\widehat{\Omat}=\Omat+\Delta\Omat$, let
$Y=\Omat z_0+\epsilon$, and estimate
$\widehat z_0=\widehat{\Omat}^{\dagger}Y$. If
$\sigma_{\min}(\Omat)>0$ and
$\lVert\Delta\Omat\rVert_2<\sigma_{\min}(\Omat)$, then
\begin{equation}
\lVert\widehat z_0-z_0\rVert_2
\le
\frac{\lVert\epsilon\rVert_2+
\lVert\Delta\Omat\rVert_2\lVert z_0\rVert_2}
{\sigma_{\min}(\Omat)-\lVert\Delta\Omat\rVert_2}.
\label{eq:operator-bound}
\end{equation}
\label{prop:perturbation}
\end{proposition}

The bound in~(\ref{eq:operator-bound}) separates measurement noise from montage/operator mismatch and motivates spectral regularization. A sealed post-primary sensitivity perturbed declared coordinates without retraining; across 50 target--condition units, reduced-operator change tracked dynamic BA change (Spearman $\rho=-0.449$; 95\% CI $-0.760$ to $-0.282$; permutation $P<0.001$; Appendix~\ref{app:coordinate-perturbation}). This characterizes the tested operators, not a universal placement tolerance. The template basis is not participant-specific localization. Proofs are in Appendix~\ref{app:proofs}.

\subsection{Static motor-spectral anchor}

For each latent component, full-window log power in $[8,13)$, $[13,30)$, and $[8,30)$~Hz gives a 24-dimensional vector. Trial $i$ receives weight $w_i=N/[2(M-1)n_{m,y_i}]$, so each source-library--class cell has equal total mass. A weighted scaler and logistic classifier are selected on source subjects and then refitted on all eligible source trials. Candidate grids and deterministic tie breaking appear in Appendix~\ref{app:implementation}.

\subsection{Stable gated state-space residual}

Using the same frequency bands, each trial is also divided into 0.5-s windows advanced at a 0.25-s stride, producing 11 vectors $x_k\in\R^{24}$. The inputs are centered within band, normalized by their RMS, and then layer-normalized before entering a 16-dimensional diagonal recurrence:
\begin{align}
c_k &= \tanh(W_cx_k+b_c), &
g_k &= \sigmoid(W_gx_k+b_g),\\
a_k &= 0.5+0.499\,\sigmoid(a_0+W_ax_k), &
h_k &= a_k\odot h_{k-1}+(1-a_k)\odot g_k\odot c_k.
\label{eq:ssm}
\end{align}
Every coordinate of $a_k$ is constrained to $(0.5,0.999)$. The residual readout combines the terminal state with the sequence-average state and applies an explicit bound,
\begin{equation}
r_{\mathrm{SSM}}=2\tanh\!\left(w^\top\left[h_K;K^{-1}\sum_{k=1}^{K}h_k\right]+b\right),
\qquad |r_{\mathrm{SSM}}|<2,
\label{eq:residual}
\end{equation}
The final sealed class probability is computed by adding this residual to the frozen static logit:
\begin{equation}
\ell=\ell_{\mathrm{static}}+r_{\mathrm{SSM}},
\qquad p(y=1\mid X)=\sigmoid(\ell).
\label{eq:final}
\end{equation}
The residual readout is initialized at zero, so the dynamic model initially returns exactly the static classifier's output. Its trainable parameters are subsequently fitted with the weighted objective
\begin{equation}
\mathcal{L}=\frac{1}{N}\sum_i w_i\left[\operatorname{BCE}(\ell_i,y_i)+10^{-2}r_i^2\right].
\label{eq:training-objective}
\end{equation}
Optimization updates the dynamic branch alone and leaves the static logit fixed. Appendix~\ref{app:implementation} specifies the source-only model selection, optimization, and refitting procedures.

\begin{figure*}[t]
  \centering
  \includegraphics[width=\textwidth]{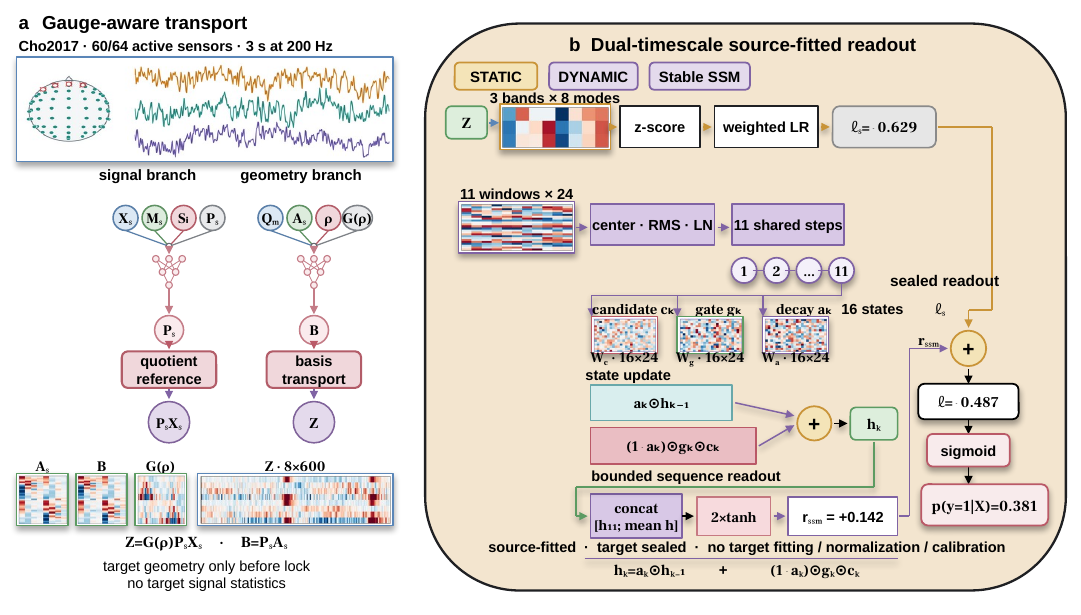}
  \caption{\textbf{Data-grounded architecture and one sealed replay.} \textbf{a}, In \texttt{Cho2017:0}, 60/64 sensors remain active over 600 samples. Separate quotient and basis-transport branches project the retained signals and geometry to the shared eight-dimensional latent; traces and $A_s$, $B$, $G(\rho)$, and $Z$ come from the label-blind replay. \textbf{b}, A frozen three-band, eight-mode static anchor is paired with an 11-window, 24-feature branch whose source-fitted SSM shares $W_c$, $W_g$, and $W_a$ and contributes a bounded $2\tanh$ residual. The trial gives $\ell_s=-0.629$, $r_{\mathrm{SSM}}=+0.142$, $\ell=-0.487$, and $p(y=1\mid X)=0.381$. Only target geometry is available before lock; all fitting, normalization, calibration, threshold selection, and model selection are source-only. Appendix~\ref{app:mechanistic-visuals} gives the full replay and stored parameters.}
  \label{fig:architecture}
\end{figure*}

\section{Experimental Design}

\subsection{Datasets and protocol}

We use five public MOABB motor-imagery collections \citep{jayaram2018moabb}: BNCI2014\_001 \citep{tangermann2012bci}, Cho2017 \citep{cho2017eeg}, Lee2019\_MI \citep{lee2019openbmi}, Shin2017A \citep{shin2017eegnirs}, and PhysionetMI \citep{schalk2004bci2000,goldberger2000physiobank}. The registered queues contain 41,342 left/right-hand trials from 253 participants (Table~\ref{tab:data}). Signals are notch filtered, band-pass filtered at 0.5--45~Hz, resampled to 200~Hz, and epoched to $[0.5,3.5)$~s. Native 22--64-channel montages are not reduced to their intersection. Fixed trial-level channel checks are detailed in Appendix~\ref{app:implementation}.

\begin{table*}[t]
\caption{Registered five-dataset evaluation and operator inventory. ``Masks'' counts retained trial-level active-channel patterns; ``inactive trials'' have at least one inactive native channel. Epochs contain 600 samples at 200~Hz over $[0.5,3.5)$~s.}
\label{tab:data}
\centering
\scriptsize
\setlength{\tabcolsep}{4pt}
\begin{tabular}{lrrrrrrr}
\toprule
Dataset & Subjects & Raw trials & Retained & Excluded & Native ch. & Masks & Inactive trials \\
\midrule
BNCI2014\_001 & 9 & 2,592 & 2,592 & 0 & 22 & 1 & 0 \\
Cho2017 & 52 & 10,520 & 10,492 & 28 & 64 & 223 & 4,710 \\
Lee2019\_MI & 54 & 21,600 & 21,600 & 0 & 62 & 6 & 34 \\
Shin2017A & 29 & 1,740 & 1,740 & 0 & 30 & 1 & 0 \\
PhysionetMI & 109 & 4,918 & 4,918 & 0 & 64 & 1 & 0 \\
\midrule
\rowcolor{GraftGray!7}
\textbf{Total} & \textbf{253} & \textbf{41,370} & \textbf{41,342} & \textbf{28} & \textbf{22--64} & \textbf{232} & \textbf{4,744} \\
\bottomrule
\end{tabular}
\end{table*}

Each fold holds out one dataset and uses four source libraries with subject-disjoint validation and seeds 2027, 3407, 4513, 5623, and 6737. Static and dynamic models share target queues. The primary endpoint is subject-mean balanced accuracy (BA); complete aggregates also report macro-F1 and AUROC, while missing metrics remain missing. Source-only calibration uses 15-bin ECE, Brier score, and negative log likelihood. Primary uncertainty is a dataset--subject hierarchical bootstrap (20,000 replicates; seed 20260829), reflecting five independent target datasets. Target effects, leave-one-target-out contrasts, and worst-domain performance accompany the aggregate; subject-paired inference is secondary.

\subsection{Comparators and protocol fairness}

The strict comparison includes BIOT, CSP--LDA, FBCSP--LDA, Riemannian--LR, EEGNet, EEGConformer, ATCNet, and CBraMod \citep{yang2023biot,blankertz2008csp,ang2008fbcsp,barachant2012riemannian,lawhern2018eegnet,song2023eegconformer,wang2025cbramod}. Frozen encoders receive source-only logistic probes and fixed geometry maps. BIOT is clean on all five targets; EEGPT and LaBraM overlap with PhysionetMI pretraining and are clean on four. CodeBrain executed all 25 cells but is provenance-ineligible on all five targets because its checkpoint origin remains unresolved; its dashes therefore denote withheld strict scores, not unexecuted cells or zero performance. Partial denominators are never rank-mixed. REVE, LUNA, and Uni-NTFM were not executed under this contract and are therefore cited as related work rather than assigned scores.

Table~\ref{tab:fairness} places target-resolved scores, execution counts, and clean eligibility denominators in one view. Target signal, empirical statistics, and labels are unavailable during fitting for every row; detailed access, input maps, parameter counts, and fitting budgets are reported in Appendix~\ref{app:fairness}.

\begin{table*}[t]
\caption{Target matrix and eligibility for the main pretrained comparison. Values are subject-mean BA (\%) over five seeds. ``Eligible mean'' weights clean targets equally; partial means are descriptive. ``Exec.'' and ``Clean'' are execution and provenance denominators. All rows are target-signal-, target-statistics-, and target-label-blind during fitting.}
\label{tab:fairness}
\centering
\scriptsize
\setlength{\tabcolsep}{4.1pt}
\begin{tabular}{lrrrrrrrr}
\toprule
Method & BNCI & Cho & Lee & PhysioNet & Shin & Eligible mean & Exec. & Clean \\
\midrule
\rowcolor{GraftBlue!8}
\method dynamic & 64.336 & 58.172 & 60.318 & 58.247 & 63.253 & \textbf{60.865} & 25/25 & 5/5 \\
\rowcolor{GraftTeal!8}
\method static & 64.923 & 56.817 & 59.448 & 56.717 & 61.126 & \underline{59.806} & 25/25 & 5/5 \\
\rowcolor{GraftOrange!10}
BIOT + LR & 56.250 & 50.086 & 52.524 & 51.741 & 52.816 & 52.684 & 25/25 & 5/5 \\
CBraMod frozen + LR & 52.392 & 50.241 & 50.341 & 50.001 & 50.724 & 50.740 & 25/25 & 5/5 \\
CBraMod source-only full FT & 50.394 & 49.925 & 50.544 & 49.908 & 47.172 & 49.589 & 25/25 & 5/5 \\
EEGPT + LR & 65.664 & 49.315 & 60.985 & --- & 60.264 & 59.057 & 25/25 & \partialden{4/5} \\
LaBraM + LR & 53.997 & 50.584 & 51.267 & --- & 52.000 & 51.962 & 25/25 & \partialden{4/5} \\
CodeBrain + LR & --- & --- & --- & --- & --- & --- & 25/25 & \provineligible{0/5} \\
\bottomrule
\end{tabular}
\end{table*}

In the equal-information analysis, FI and SSI receive the same active set as \method, map each mask to 17 fixed channels, and use source-fitted tangent features or EEGNet \citep{mellot2024physics,gramfort2013mne,ledoit2004covariance}. All 60 cells completed fit, prediction, and scoring. Supplementary analyses cover calibration, stress, reference, stability, and known-source operators; structural gaps remain N/A. Dreyer2023A \citep{dreyer2023large} is a no-tuning same-paradigm evaluation with partial-confirmation status. Appendices~\ref{app:implementation}--\ref{app:audit} give definitions and verification order.

The post-hoc exact same-parent ladder replaces the physical leadfield with one of four operators: a constant-free real spherical-harmonic coordinate basis, a random operator with the physical singular spectrum, a fixed derangement of electrode rows, or a visual-cortex leadfield from the same template anatomy. Each variant is refitted on the same source split under the inherited FULL budget; target geometry, active-reference quotient, ridge rule, dynamic head, seeds, features, and evaluation queues remain fixed. All 75 new fits were frozen before any target evaluation. Seeds are averaged within target--subject, subjects within target, and the five targets equally; intervals use 20,000 target-stratified subject-bootstrap replicates with Holm correction across the four controls. These post-hoc comparisons are not independently confirmatory.

\section{Results}

\subsection{Strict five-target comparison}

\begin{figure*}[t]
  \centering
  \includegraphics[width=0.99\textwidth]{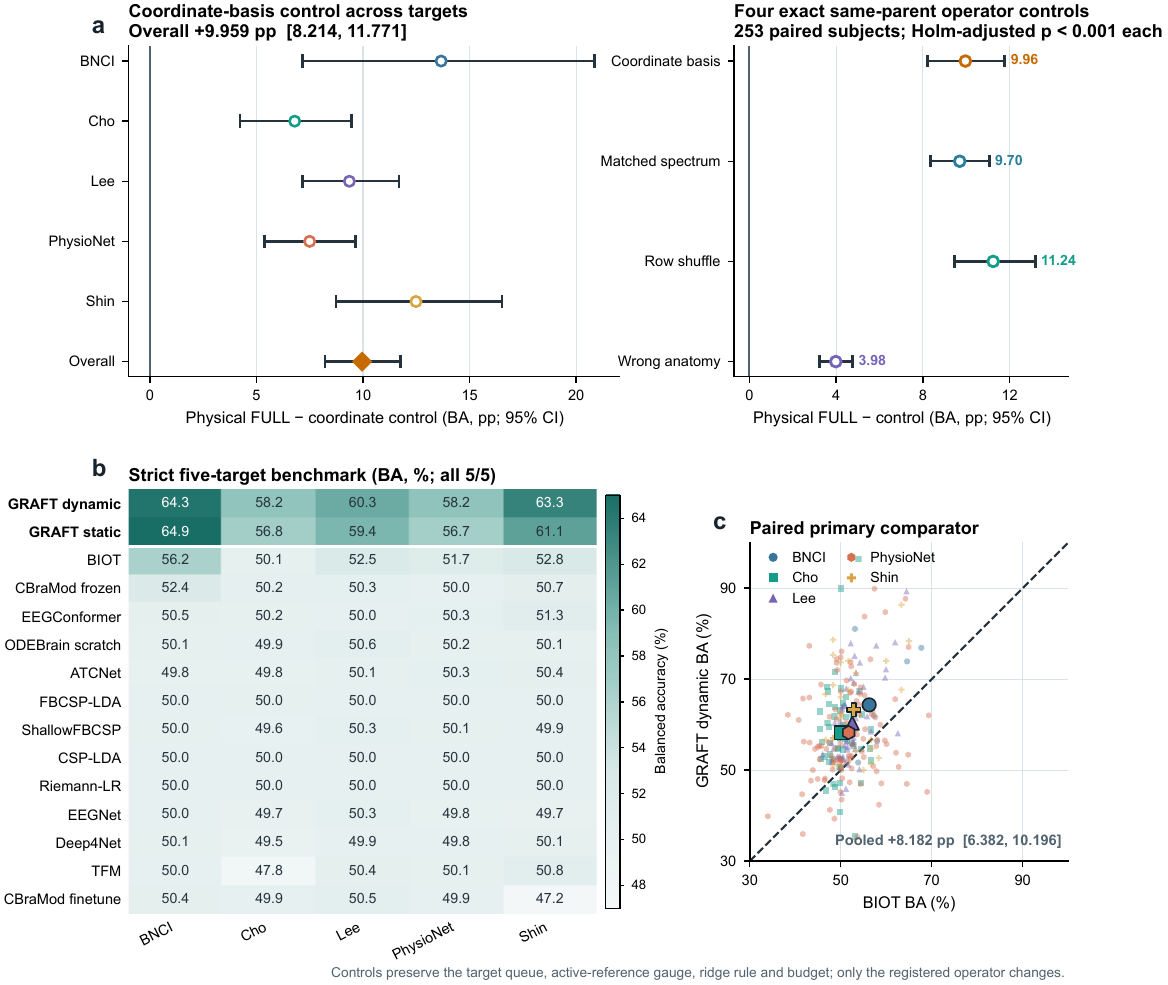}
  \caption{\textbf{Physical-leadfield mechanism controls and strict cross-dataset performance.} \textbf{a}, Left: physical FULL minus the layout-aware coordinate basis by target and overall. Right: equal-target effects for all four exact same-parent controls; intervals use 20,000 target-stratified paired-subject bootstrap replicates after averaging five seeds within subject, and every Holm-adjusted $P<0.001$. All controls retain the split, features, head, ridge rule, budget, and target queues; the replacement operator is the only intended change. \textbf{b}, Strict-clean 5/5 balanced-accuracy matrix for all 15 complete methods. \textbf{c}, Dynamic \method versus BIOT for the same 253 held-out subjects. Diamonds denote target means and the dashed line denotes identity. The operator ladder is post-hoc and not independent external validation.}
  \label{fig:main-results}
\end{figure*}

Dynamic \method achieved 60.865\% equal-target BA, static \method 59.806\%, and BIOT 52.684\% (Figure~\ref{fig:main-results}c). Dynamic exceeded BIOT by $+8.182$~pp (hierarchical 95\% interval, $+6.382$ to $+10.196$); all five target effects were positive ($+6.506$ to $+10.437$~pp), and leave-one-target-out estimates were 7.618--8.600~pp. Cho was the worst target for both dynamic \method (58.172\%) and BIOT (50.086\%).

The exact same-parent ladder separated FULL from every replacement operator (Figure~\ref{fig:main-results}a). FULL exceeded the coordinate basis by $+9.959$~pp (95\% CI $+8.214$ to $+11.771$), the matched-spectrum random operator by $+9.698$~pp ($+8.348$ to $+11.052$), the row-shuffled operator by $+11.241$~pp ($+9.450$ to $+13.174$), and the wrong-anatomy leadfield by $+3.985$~pp ($+3.231$ to $+4.752$). All four target-stratified paired-subject tests had two-sided bootstrap $P<0.001$ and remained below $0.001$ after Holm correction. The coordinate effect was positive in every target ($+6.79$ to $+13.67$~pp). These controls distinguish the tested motor leadfield from geometry-only, spectrum-preserving, correspondence-breaking, and anatomically mismatched alternatives; they do not establish superiority over every possible physical or learned operator.

The SSM is secondary to cross-montage harmonization: dynamic minus static was $+1.059$~pp (hierarchical 95\% interval, $+0.013$ to $+1.892$; Table~\ref{tab:lodo-sensitivity}). Four target effects were positive, whereas BNCI was $-0.586$~pp ($-2.423$ to $+1.497$); leave-one-target-out estimates were 0.792--1.470~pp. The tighter subject-paired interval ($+0.584$ to $+1.569$~pp) is within-panel evidence, not the principal cross-dataset interval.

In the stricter source-only-geometry sensitivity, all 25 fits were frozen before the target montage was opened. Dynamic and static BA were 60.460\% and 59.713\%, respectively; dynamic was 0.406~pp below its canonical result and exceeded static by $+0.747$~pp. The held-out montage is still required after sealing to instantiate the fixed prediction operator, so the method is not geometry-blind. Per-target values and the independent audit are in Appendix~\ref{app:strict-geometry}.

\subsection{Coverage, fairness, and external boundary}

BIOT is the best-performing clean pretrained comparator with a 5/5 denominator and is therefore the primary foundation-model comparison. EEGPT and LaBraM executed 25/25 target--seed cells but are strict-clean on 4/5 targets because PhysionetMI overlaps with pretraining. CodeBrain also executed 25/25 cells but is provenance-ineligible on 0/5 targets; no CodeBrain score enters a strict comparison. On the same four clean targets available to EEGPT (BNCI, Cho, Lee, and Shin), dynamic \method averaged 61.520\% BA and EEGPT 59.057\%, a descriptive difference of $+2.463$~pp. The target effects were $-1.328$, $+8.857$, $-0.667$, and $+2.989$~pp, respectively. This 4/5 comparison is not inserted into the complete 5/5 ranking. Figure~\ref{fig:benchmark-health} preserves the full matrix and all eligibility denominators.

A read-only post-score diagnostic checked source validation, target positive rates, prediction entropy, target--seed coverage, and single-class counts without training or parameter updates. Dynamic and static \method met the prespecified healthy-source rule and had no single-class target cells. BIOT was nondegenerate but had marginal source validation (mean BA 53.6\%); several other baselines were already near chance on source validation, and the classical CSP/FBCSP/Riemann rows had systematic single-class target predictions with source validation unavailable. No direct implementation defect was confirmed. Accordingly, weak target results are not attributed uniformly to domain shift, and collapsed or source-weak baselines are not used as mechanistic evidence for \method; Appendix~\ref{app:baseline-health} reports the complete diagnostic.

Supplementary analyses retain the original denominators. Under the registered few-shot procedure, revealing 1\%, 5\%, or 10\% target labels and fine-tuning the full GRAFT head gave 61.323\%, 61.477\%, and 61.722\% BA, respectively, compared with 60.865\% in the zero-target protocol. These small increments contextualize the absolute zero-target BA under this adaptation rule; they are not a fully supervised oracle or a performance ceiling. In the separate 60-participant Dreyer2023A evaluation, dynamic versus static was $+0.57$~pp [$-0.08,+1.23$] (inconclusive), whereas physics versus no-physics was $+4.03$~pp [$+2.34,+5.77$] (positive). The preregistered decision is \emph{partial confirmation}, not unrestricted external validation.

The post-primary coordinate sensitivity refitted no model. Across 50 target--condition units, reduced-operator change tracked dynamic BA change ($\rho=-0.449$; 95\% CI $-0.760$ to $-0.282$; permutation $P<0.001$); at 20~mm, equal-target BA changed by $-1.064$~pp for tangent and $-2.220$~pp for rigid errors. Responses were heterogeneous (Figure~\ref{fig:coordinate-perturbation}).

\section{Discussion and Limitations}

The central finding concerns the observation operator rather than the temporal head. Without target signals, labels, or statistics during fitting, the native-montage model exceeded the strongest clean 5/5 pretrained comparator by 8.18~pp in every target. Under the same parent and budget, replacing the motor leadfield with four distinct controls reduced BA by 3.98--11.24~pp. The smaller but still positive wrong-anatomy contrast shows that an anatomically structured template is not interchangeable with the task-matched motor field; the larger coordinate, spectrum-matched, and row-shuffled contrasts additionally reject explanations based only on dimension, singular spectrum, or electrode-wise response magnitude. The ladder remains post-hoc and operator-specific rather than proving that physics is universally optimal. Propositions~\ref{prop:identifiability} and~\ref{prop:perturbation} state the corresponding observability and conditioning limits.

The SSM contribution was modest and heterogeneous ($+1.06$~pp overall; four positive targets; $-0.59$~pp on BNCI; inconclusive external contrast), supporting an incremental interpretation. The row-shuffled and spectrum-matched controls in Appendix~\ref{app:known-source} evaluate inverse recovery with synthetic known sources; they are not a real-EEG same-parent classification ladder. Participant-specific source recovery lies beyond scope. After the primary scores were fixed, three of four matched FI/SSI intervals excluded zero; FI+tangent crossed zero, so this supplementary evidence remains contrast-specific rather than supporting a general superiority claim.

Calibration lowered probability error while leaving discrimination unchanged, and performance losses under stress reached $-8.93$~pp. In the separately registered legacy three-target stability diagnostics, \method showed greater seed-to-seed dispersion than the coordinate and no-physics comparators. Those adverse dispersion results are distinct from the new five-target exact same-parent performance control. Failed conditions also outnumbered recoverable ones in the known-source tests. Claims are consequently confined to the operating range characterized by these evaluations, rather than extended to unrestricted inverse recovery.

The independent generalization unit is the held-out dataset ($n=5$), not the 253 participants, and dataset remains confounded with device and protocol. The primary task is binary hand motor imagery; non-motor paradigms were not tested. Blindness applies to target statistics, not declared target coordinates, and the source-only-geometry experiment is sensitivity analysis rather than a replacement for the fixed primary result. The fsaverage basis is a common coordinate system, not participant-specific localization, so causal and clinical interpretations remain outside scope. Coordinate sensitivity is now characterized only for two synthetic error families on the five declared montages; it neither identifies a clinical placement threshold nor covers arbitrary cap deformation or anatomy. Matched FI/SSI and other post-score diagnostics are supplementary; arbitrary montages, streaming, and clinical deployment remain outside the evidence boundary.

\section{Conclusion}

Across five leave-one-dataset-out targets, \method operated on native montages and trial-level active masks without using target signals, labels, or fitting statistics. Dynamic \method achieved 60.865\% equal-target BA, compared with 52.684\% for strict BIOT. Four exact same-parent replacements reduced BA by 3.985--11.241~pp, supporting a material contribution from the task-matched physical observation operator within this parent model. This remains operator-specific evidence from five binary motor-imagery domains, not a universal claim over geometry models or EEG paradigms. The bounded SSM supplied a smaller, target-dependent increment.

\noindent\textbf{Reproducibility.} Fitting, label-blind prediction, and scoring are process-separated. Immutable manifests bind queues, masks, checkpoints, predictions, and result tables; complete gates and source-data hashes appear in Appendices~\ref{app:audit}--\ref{app:source-manifest}. The reproducibility package retains figure source data and executed-cell ledgers for every denominator.

\noindent\textbf{Ethics.} The study uses public, de-identified EEG data and recruited no participants. No re-identification was attempted; original consent, license, and access conditions continue to govern each dataset. The work evaluates decoding methodology, not diagnosis or treatment.

\noindent\textbf{AI use.} A large language model was used solely for grammar correction and language polishing. The study design, code development, experiments and analyses were completed by the authors, who take full responsibility for the manuscript.

\clearpage
\bibliography{references}
\bibliographystyle{iclr2027_conference}

\appendix
\setcounter{figure}{0}
\renewcommand{\thefigure}{S\arabic{figure}}
\renewcommand{\theHfigure}{app.\arabic{figure}}
\setcounter{table}{0}
\renewcommand{\thetable}{S\arabic{table}}
\renewcommand{\theHtable}{app.\arabic{table}}

\section{Proofs of the Restricted Theory Results}
\label{app:proofs}

\begin{proof}[Proof of Proposition~\ref{prop:identifiability}]
With the noiseless measurements stacked, $Y=\Omat_{\mathcal M,T}z_0$. If $z_0$ and $z_0'$ yield the same $Y$, then $\Omat_{\mathcal M,T}(z_0-z_0')=0$; full column rank therefore gives $z_0=z_0'$. Identifiability is nevertheless limited to a latent-coordinate class. For any invertible $S$, let $z_t'=Sz_t$, $F'=SFS^{-1}$, and $A_m'=A_mS^{-1}$. Because $z_{t+1}'=F'z_t'$ and $A_m'z_t'=A_mz_t$, this transformation leaves every observation unchanged.
\end{proof}

\begin{proof}[Proof of Proposition~\ref{prop:perturbation}]
Weyl's inequality and the assumed perturbation bound give
$\sigma_{\min}(\widehat\Omat)\ge
\sigma_{\min}(\Omat)-\lVert\Delta\Omat\rVert_2>0$.
Hence $\widehat\Omat$ retains full column rank, and
\begin{align*}
\widehat z_0-z_0
&=\widehat\Omat^\dagger
(\Omat z_0+\epsilon)-z_0 \\
&=\widehat\Omat^\dagger
(\epsilon-\Delta\Omat z_0).
\end{align*}
Since
$\lVert\widehat\Omat^\dagger\rVert_2=
1/\sigma_{\min}(\widehat\Omat)$, the triangle inequality gives the bound in~(\ref{eq:operator-bound}).
\end{proof}

The scope of these results is deliberately narrow: both propositions apply to a fixed, finite-dimensional linearization. Neither guarantees unique recovery for the nonlinear EEG inverse problem nor identifies participant-specific cortical generators.

\FloatBarrier
\section{Dataset and Observation-Operator Inventory}
\label{app:operator-inventory}

Table~\ref{tab:operator-inventory} expands the acquisition summary in Table~\ref{tab:data}. The retained queue is not a rectangular common-channel tensor: Cho2017 alone contributes 223 distinct active masks, and 4,710 of its retained trials have at least one inactive native channel. The operator cache is therefore indexed by the exact retained mask rather than by dataset alone. The inventory is computed from the immutable formal queues; it is descriptive and does not reopen target outcomes.

\begin{table*}[h]
\centering
\caption{Detailed immutable queue and operator inventory. ``Raw'' is the prepared H5 queue before the fixed trial gate. ``Patterns'' is the number of distinct retained active-channel masks.}
\label{tab:operator-inventory}
\small
\setlength{\tabcolsep}{4pt}
\begin{tabular}{lrrrrrrrrr}
\toprule
Dataset & Subj. & Raw & Retained & Excl. & Ch. & Samples & Hz & Patterns & Inactive trials \\
\midrule
BNCI2014\_001 & 9 & 2,592 & 2,592 & 0 & 22 & 600 & 200 & 1 & 0 \\
Cho2017 & 52 & 10,520 & 10,492 & 28 & 64 & 600 & 200 & 223 & 4,710 \\
Lee2019\_MI & 54 & 21,600 & 21,600 & 0 & 62 & 600 & 200 & 6 & 34 \\
PhysionetMI & 109 & 4,918 & 4,918 & 0 & 64 & 600 & 200 & 1 & 0 \\
Shin2017A & 29 & 1,740 & 1,740 & 0 & 30 & 600 & 200 & 1 & 0 \\
\midrule
Total & 253 & 41,370 & 41,342 & 28 & --- & --- & --- & 232 & 4,744 \\
\bottomrule
\end{tabular}
\end{table*}

The 28 exclusions all occur in Cho2017 under the same prelocked amplitude, finiteness, and active-channel-count rules used for every dataset. BNCI2014\_001, PhysionetMI, and Shin2017A each retain a single all-active pattern; Lee2019\_MI retains six patterns. These counts explain why a single fixed pseudo-inverse per dataset would be scientifically incorrect even though the nominal montage is fixed.

\FloatBarrier

\section{Strict Source-Only Geometry-Basis Sensitivity}
\label{app:strict-geometry}

For target fold $m^\star$, this sensitivity constructs the hemisphere-blocked $d=8$ basis from the four source geometry operators only. It selects every hyperparameter from the same four source datasets and seals all 25 target--seed fits before opening any target geometry or target H5 file. The target montage is then used once with the frozen fold-specific basis to construct the prediction operator; target signals and labels remain separated into inference and scoring processes. This test changes only the geometry-access boundary and does not alter or supersede the canonical sealed results.

\begin{table*}[h]
\centering
\caption{Access-boundary sensitivity: canonical target-geometry construction versus a strict source-only geometry basis. Values are subject-mean balanced accuracy averaged over five seeds. ``Change'' is source-only dynamic minus canonical dynamic; it isolates the cost of withholding the target montage until after the source fits are sealed.}
\label{tab:strict-geometry}
\scriptsize
\setlength{\tabcolsep}{4pt}
\begin{tabular}{lrrrrr}
\toprule
Target ($n$) & Canon. dyn. & Src.-only dyn. & Change (pp) & Src.-only stat. & Dyn.--stat. (pp) \\
\midrule
BNCI2014-001 (9) & 64.336 & 62.423 & $-1.913$ & 64.668 & $-2.245$ \\
Cho2017 (52)      & 58.172 & 58.141 & $-0.031$ & 56.868 & $+1.273$ \\
Lee2019-MI (54)   & 60.318 & 60.392 & $+0.074$ & 59.361 & $+1.031$ \\
PhysionetMI (109) & 58.247 & 58.020 & $-0.227$ & 56.724 & $+1.296$ \\
Shin2017A (29)    & 63.253 & 63.322 & $+0.069$ & 60.943 & $+2.379$ \\
\midrule
Equal-target mean (253) & 60.865 & 60.460 & $-0.406$ & 59.713 & $+0.747$ \\
\bottomrule
\end{tabular}
\end{table*}

The source-only experiment completed all 25 fits and all 25 frozen evaluations. Replaying the predictions in a fresh process returned \texttt{PASS} ($P0=P1=0$), and the run ended with status \texttt{completed} and exit code~0. The global fit barrier had SHA-256 \hashsplit{fff8c450}{91a74a19}{5a4d9ff1}{dfd2c854}{64739bea}{699937be}{31de8451}{9a847b9b}; the final audit seal had SHA-256 \hashsplit{ff7fbd82}{098e0602}{02fcd236}{f1ccb155}{355cb785}{4ab4a025}{e81754fc}{715953b0}. The original 60-cell score anchor did not change. Source-only geometry reduced the canonical dynamic mean by 0.406~pp, much less than the canonical model's 8.182-pp advantage over BIOT. BNCI nevertheless retained a negative dynamic--static contrast, so the SSM effect should be read as target-dependent.

\FloatBarrier
\section{Complete Strict Five-Target Discrimination Summary}
\label{app:strict-metrics}

Table~\ref{tab:strict-metrics} reports every method with a clean 5/5 execution denominator. All entries use the same equal-target aggregation: seed means are formed within target and targets then receive equal weight. Macro-F1 or AUROC is shown only when that metric is present for all five targets in the sealed aggregate. The dashes for CBraMod frozen, ATCNet, ShallowFBCSPNet, and Deep4Net are retained missing values, not failed or zero-valued metrics.

\begin{table*}[h]
\centering
\caption{Complete strict five-target discrimination summary. Values are percentages. All 15 methods completed 25/25 target--seed cells and are strict-eligible on 5/5 targets. Missing secondary metrics are not imputed.}
\label{tab:strict-metrics}
\small
\begin{tabular}{lrrrl}
\toprule
Method & BA & Macro-F1 & AUROC & Metric availability \\
\midrule
\rowcolor{GraftBlue!8}
\method dynamic & \textbf{60.865} & \textbf{58.886} & \textbf{66.447} & all three sealed \\
\rowcolor{GraftTeal!8}
\method static & \underline{59.806} & \underline{58.015} & \underline{64.588} & all three sealed \\
\rowcolor{GraftOrange!10}
BIOT + LR & 52.684 & 46.954 & 56.131 & all three sealed \\
CBraMod frozen + LR & 50.740 & --- & --- & BA only \\
EEGConformer & 50.448 & 44.322 & 50.808 & all three sealed \\
ODEBrain (scratch) & 50.174 & 39.996 & 51.060 & all three sealed \\
ATCNet & 50.074 & --- & --- & BA only \\
FBCSP--LDA & 50.005 & 33.400 & 50.302 & all three sealed \\
ShallowFBCSPNet & 50.004 & --- & --- & BA only \\
CSP--LDA & 50.000 & 33.369 & 50.060 & all three sealed \\
Riemann--LR & 49.996 & 33.367 & 50.034 & all three sealed \\
EEGNet & 49.901 & 43.004 & 49.784 & all three sealed \\
Deep4Net & 49.895 & --- & --- & BA only \\
TFM & 49.814 & 43.458 & 49.449 & all three sealed \\
CBraMod source-only full FT & 49.589 & 40.214 & 49.467 & all three sealed \\
\bottomrule
\end{tabular}
\end{table*}

Table~\ref{tab:targetwise-strict-ba} exposes the target matrix behind the equal-target means. This prevents a high-performing large dataset from dominating the ranking and makes the worst-domain behavior visible. The best strict comparator is BIOT on every target, but its target values remain below dynamic \method by 6.506--10.437~pp.

\begin{table*}[h]
\centering
\caption{Target-resolved subject-mean balanced accuracy for every strict-clean 5/5 method. Values are percentages after averaging the five registered seeds within target. The last column weights targets equally.}
\label{tab:targetwise-strict-ba}
\scriptsize
\setlength{\tabcolsep}{4pt}
\begin{tabular}{lrrrrrr}
\toprule
Method & BNCI & Cho & Lee & PhysioNet & Shin & Equal-target \\
\midrule
\rowcolor{GraftBlue!8}
\method dynamic & \textbf{64.336} & \textbf{58.172} & \textbf{60.318} & \textbf{58.247} & \textbf{63.253} & \textbf{60.865} \\
\rowcolor{GraftTeal!8}
\method static & \underline{64.923} & \underline{56.817} & \underline{59.448} & \underline{56.717} & \underline{61.126} & \underline{59.806} \\
\rowcolor{GraftOrange!10}
BIOT + LR & 56.250 & 50.086 & 52.524 & 51.741 & 52.816 & 52.684 \\
CBraMod frozen + LR & 52.392 & 50.241 & 50.341 & 50.001 & 50.724 & 50.740 \\
EEGConformer & 50.455 & 50.183 & 49.958 & 50.321 & 51.322 & 50.448 \\
ODEBrain (scratch) & 50.085 & 49.871 & 50.621 & 50.210 & 50.080 & 50.174 \\
ATCNet & 49.792 & 49.750 & 50.064 & 50.350 & 50.414 & 50.074 \\
FBCSP--LDA & 50.000 & 50.027 & 50.000 & 50.000 & 50.000 & 50.005 \\
ShallowFBCSPNet & 50.031 & 49.630 & 50.334 & 50.083 & 49.943 & 50.004 \\
CSP--LDA & 50.000 & 50.000 & 50.000 & 50.000 & 50.000 & 50.000 \\
Riemann--LR & 50.000 & 49.980 & 50.000 & 50.000 & 50.000 & 49.996 \\
EEGNet & 49.977 & 49.697 & 50.345 & 49.785 & 49.701 & 49.901 \\
Deep4Net & 50.108 & 49.526 & 49.933 & 49.836 & 50.069 & 49.895 \\
TFM & 49.977 & 47.816 & 50.419 & 50.088 & 50.770 & 49.814 \\
CBraMod source-only full FT & 50.394 & 49.925 & 50.544 & 49.908 & 47.172 & 49.589 \\
\bottomrule
\end{tabular}
\end{table*}

\begin{table*}[h]
\centering
\caption{Executed but strict-incomplete pretrained encoders. A numeric cell is shown only when the target is clean under the checkpoint-provenance review. The available-target mean is descriptive and is not inserted into Table~\ref{tab:targetwise-strict-ba}.}
\label{tab:partial-denominators}
\small
\setlength{\tabcolsep}{4pt}
\begin{tabularx}{\textwidth}{lrrrrrrr>{\raggedright\arraybackslash}X}
\toprule
Method & BNCI & Cho & Lee & PhysioNet & Shin & Available mean & Strict & Exclusion \\
\midrule
EEGPT + LR & 65.664 & 49.315 & 60.985 & --- & 60.264 & 59.057 & \partialden{4/5} & overlap \\
LaBraM + LR & 53.997 & 50.584 & 51.267 & --- & 52.000 & 51.962 & \partialden{4/5} & overlap \\
CodeBrain + LR & --- & --- & --- & --- & --- & --- & \provineligible{0/5} & unresolved \\
\bottomrule
\end{tabularx}
\end{table*}

\FloatBarrier
\section{Post-Seal Baseline Health Diagnostics}
\label{app:baseline-health}

The health audit is a read-only diagnostic over frozen probabilities and source-validation records; it started no training or evaluation job and changed no registered result. Table~\ref{tab:baseline-health} reports the quantities requested before interpreting near-chance target scores. Source validation is unavailable for the three classical pipelines, and CodeBrain has no health row bound to a strict-clean checkpoint identity. These absences remain N/A. A single-class count means that the thresholded predictions in a target--seed cell contain only one label; it is not inferred from mean probability entropy alone.

\begin{table*}[h]
\centering
\caption{Post-seal implementation and prediction-health audit. Source BA is the mean source-validation balanced accuracy (\%). ``Positive-rate range'' is the minimum--maximum target positive prediction rate across eligible target--seed cells (\%). $H_{\mathrm{hard}}$ and $H_{\mathrm{prob}}$ are mean hard-label and mean probability entropy in bits. ``Single'' counts single-class target cells; ``coverage'' is completed/eligible health-audit cells. N/A values are not imputed.}
\label{tab:baseline-health}
\scriptsize
\setlength{\tabcolsep}{3.2pt}
\resizebox{\textwidth}{!}{%
\begin{tabular}{lrrrrrr}
\toprule
Method & Source BA & Positive-rate range & $H_{\mathrm{hard}}$ & $H_{\mathrm{prob}}$ & Single & Coverage \\
\midrule
\rowcolor{GraftBlue!8}
\method dynamic & 60.8 & 26.4--56.2 & 0.974 & 0.902 & 0 & 25/25 \\
\rowcolor{GraftTeal!8}
\method static & 58.6 & 39.8--55.4 & 0.993 & 0.948 & 0 & 25/25 \\
\rowcolor{GraftOrange!10}
BIOT & 53.6 & 26.9--76.8 & 0.870 & 0.859 & 0 & 25/25 \\
EEGPT + LR & 61.2 & 13.3--54.4 & 0.870 & 0.830 & 0 & 20/20 \\
LaBraM + LR & 52.4 & 37.0--75.6 & 0.929 & 0.978 & 0 & 20/20 \\
CodeBrain + LR & N/A & N/A & N/A & N/A & N/A & 25/25$^\dagger$ \\
CBraMod frozen + LR & 51.4 & 0.0--90.5 & 0.667 & 0.993 & 2 & 25/25 \\
CBraMod source-only full FT & 52.8 & 0.0--100.0 & 0.528 & 0.981 & 2 & 25/25 \\
EEGConformer & 51.2 & 2.5--98.1 & 0.728 & 0.993 & 0 & 25/25 \\
EEGNet & 51.2 & 0.2--97.6 & 0.637 & 0.965 & 0 & 25/25 \\
ATCNet & 51.3 & 0.2--99.9 & 0.570 & 0.973 & 0 & 25/25 \\
ShallowFBCSPNet & 50.8 & 0.0--100.0 & 0.390 & 0.881 & 3 & 25/25 \\
Deep4Net & 50.9 & 0.0--100.0 & 0.414 & 0.916 & 4 & 25/25 \\
TFM & 50.9 & 0.0--49.9 & 0.751 & 0.991 & 0 & 25/25 \\
ODEBrain (scratch) & 51.3 & 0.0--95.4 & 0.534 & 0.967 & 2 & 25/25 \\
CSP--LDA & N/A & 0.0--100.0 & 0.000 & 1.000 & 25 & 25/25 \\
FBCSP--LDA & N/A & 0.0--99.9 & 0.003 & 1.000 & 20 & 25/25 \\
Riemann--LR & N/A & 0.0--100.0 & 0.001 & 1.000 & 20 & 25/25 \\
Separate graft parent FULL & 51.0 & 0.0--100.0 & 0.727 & 0.999 & 2 & 25/25 \\
\bottomrule
\end{tabular}}

\vspace{2pt}
\raggedright\scriptsize $^\dagger$CodeBrain executed 25/25 cells in the coverage ledger, but unresolved checkpoint provenance prevented a strict-clean model identity and a bound health summary; no health conclusion is assigned.
\end{table*}

The independent implementation-health audit returned PASS ($P0=P1=0$) for input hashes, runner identities, coverage, probability validity, train/eval-mode checks, and artifact consistency. It found no confirmed code defect, but explicitly did not convert near-chance source validation into healthy baselines. The prespecified domain-shift attribution rule was met in zero cells. Thus, the table supports health-aware interpretation and disclosure, not post hoc removal or reranking.

The locked subject-paired analysis predates the primary dataset--subject hierarchical bootstrap. Table~\ref{tab:secondary-inference} retains it as secondary within-panel evidence because it asks whether the same 253 held-out participants tend to favor dynamic \method. The new exact same-parent row includes all five targets. The legacy no-physics and wrong-operator diagnostics include only their three originally registered targets and are labeled accordingly. Every Holm-adjusted Wilcoxon decision in the underlying paired tables rejects at 0.05, but those subject-level decisions are not interpreted as evidence from 253 independent datasets.

\begin{table*}[h]
\centering
\caption{Secondary subject-paired physical-dynamic-minus-comparator effects in percentage points. Each entry is effect [95\% paired subject-bootstrap interval], based on 10,000 replicates. The exact same-parent row uses its registered target-stratified subject bootstrap; the primary dynamic--BIOT and dynamic--static contrasts in the main text instead use dataset--subject hierarchical resampling.}
\label{tab:secondary-inference}
\scriptsize
\setlength{\tabcolsep}{3.2pt}
\resizebox{\textwidth}{!}{%
\begin{tabular}{lccccl}
\toprule
Comparator & Balanced accuracy & Macro-F1 & AUROC & Targets / subjects & Holm decision \\
\midrule
Exact same-parent layout-aware coordinate control & $+9.96$ [$+8.20,+11.75$] & $+12.31$ [$+10.22,+14.39$] & $+14.88$ [$+11.95,+17.90$] & 5 / 253 & reject \\
\method static & $+1.06$ [$+0.58,+1.57$] & $+0.87$ [$+0.35,+1.41$] & $+1.86$ [$+1.40,+2.30$] & 5 / 253 & reject \\
CSP--LDA & $+10.87$ [$+9.22,+12.59$] & $+25.52$ [$+23.58,+27.47$] & $+16.39$ [$+14.01,+18.90$] & 5 / 253 & reject \\
FBCSP--LDA & $+10.86$ [$+9.18,+12.61$] & $+25.49$ [$+23.55,+27.45$] & $+16.14$ [$+13.80,+18.67$] & 5 / 253 & reject \\
Riemann--LR & $+10.87$ [$+9.25,+12.65$] & $+25.52$ [$+23.59,+27.49$] & $+16.41$ [$+13.95,+18.93$] & 5 / 253 & reject \\
EEGNet & $+10.96$ [$+9.28,+12.67$] & $+15.88$ [$+13.95,+17.86$] & $+16.66$ [$+14.24,+19.21$] & 5 / 253 & reject \\
EEGConformer & $+10.42$ [$+8.67,+12.23$] & $+14.56$ [$+12.56,+16.60$] & $+15.64$ [$+13.14,+18.35$] & 5 / 253 & reject \\
Legacy no-physics dynamic (3-target) & $+9.78$ [$+8.17,+11.49$] & $+11.05$ [$+9.20,+12.96$] & $+14.82$ [$+12.62,+17.14$] & 3 / 192 & reject \\
Legacy no-physics static (3-target) & $+7.96$ [$+6.35,+9.65$] & $+9.39$ [$+7.58,+11.20$] & $+12.44$ [$+10.10,+14.91$] & 3 / 192 & reject \\
Legacy wrong-operator dynamic (3-target) & $+8.69$ [$+7.26,+10.14$] & $+15.79$ [$+14.25,+17.37$] & $+13.71$ [$+11.37,+16.00$] & 3 / 192 & reject \\
Legacy wrong-operator static (3-target) & $+8.48$ [$+6.91,+10.04$] & $+16.20$ [$+14.45,+17.99$] & $+11.54$ [$+9.33,+13.81$] & 3 / 192 & reject \\
\bottomrule
\end{tabular}}
\end{table*}

\FloatBarrier
\section{Registered Cell Reproducibility and Subject Heterogeneity}
\label{app:cell-reproducibility}

The primary summaries average five registered seeds within each target before giving equal weight to the five targets. Table~\ref{tab:registered-cells} exposes the corresponding 25 target--seed cells for both locked \method branches. These rows are a pivot of the sealed \texttt{cell\_metrics.csv}; no model was refitted, no prediction was rescored, and no seed was removed after inspection. The seed cells show computational repeatability within a target, but they are nested observations rather than 25 independent target domains.

\begin{table*}[h]
\centering
\caption{Registered target--seed cell metrics for dynamic (D) and static (S) \method. BA, macro-F1, and AUROC are percentages; $\Delta$BA is D minus S in percentage points. Every target contains the same five preregistered seeds.}
\label{tab:registered-cells}
\scriptsize
\setlength{\tabcolsep}{3.0pt}
\renewcommand{\arraystretch}{0.94}
\begin{tabular}{lrrrrrrrr}
\toprule
Target & Seed & BA D & BA S & $\Delta$BA & F1 D & F1 S & AUROC D & AUROC S \\
\midrule
BNCI & 2027 & 63.310 & 64.931 & $-1.620$ & 61.234 & 63.947 & 69.688 & 70.054 \\
BNCI & 3407 & 62.693 & 64.429 & $-1.736$ & 60.029 & 63.240 & 70.720 & 70.224 \\
BNCI & 4513 & 64.738 & 64.660 & $+0.077$ & 63.406 & 63.684 & 69.076 & 69.882 \\
BNCI & 5623 & 62.384 & 64.429 & $-2.045$ & 59.294 & 63.240 & 71.056 & 70.224 \\
BNCI & 6737 & 68.557 & 66.165 & $+2.392$ & 67.474 & 65.274 & 74.157 & 71.401 \\
\midrule
Cho & 2027 & 58.073 & 56.794 & $+1.280$ & 55.579 & 54.249 & 62.714 & 60.348 \\
Cho & 3407 & 58.185 & 56.874 & $+1.311$ & 55.905 & 54.439 & 62.770 & 60.482 \\
Cho & 4513 & 58.082 & 56.794 & $+1.288$ & 55.602 & 54.249 & 62.744 & 60.348 \\
Cho & 5623 & 58.018 & 56.819 & $+1.199$ & 55.817 & 54.417 & 62.975 & 60.513 \\
Cho & 6737 & 58.499 & 56.806 & $+1.694$ & 56.528 & 54.638 & 62.983 & 60.478 \\
\midrule
Lee & 2027 & 60.435 & 59.500 & $+0.935$ & 59.019 & 57.714 & 65.513 & 64.587 \\
Lee & 3407 & 60.310 & 59.519 & $+0.792$ & 58.331 & 57.800 & 66.001 & 64.544 \\
Lee & 4513 & 60.468 & 59.574 & $+0.894$ & 58.832 & 58.030 & 65.929 & 64.443 \\
Lee & 5623 & 60.185 & 59.514 & $+0.671$ & 58.081 & 57.792 & 65.868 & 64.541 \\
Lee & 6737 & 60.190 & 59.134 & $+1.056$ & 58.263 & 57.546 & 65.645 & 63.865 \\
\midrule
PhysioNet & 2027 & 58.370 & 56.639 & $+1.731$ & 56.145 & 54.206 & 62.993 & 60.794 \\
PhysioNet & 3407 & 58.068 & 56.852 & $+1.216$ & 56.135 & 54.276 & 61.985 & 60.585 \\
PhysioNet & 4513 & 58.741 & 56.620 & $+2.121$ & 56.446 & 54.150 & 63.221 & 60.804 \\
PhysioNet & 5623 & 57.862 & 56.852 & $+1.010$ & 55.131 & 54.276 & 62.650 & 60.585 \\
PhysioNet & 6737 & 58.194 & 56.620 & $+1.574$ & 55.765 & 54.150 & 63.087 & 60.804 \\
\midrule
Shin & 2027 & 62.816 & 61.264 & $+1.552$ & 61.393 & 59.727 & 69.008 & 67.303 \\
Shin & 3407 & 63.563 & 61.264 & $+2.299$ & 62.116 & 59.727 & 70.368 & 67.291 \\
Shin & 4513 & 63.506 & 61.264 & $+2.241$ & 62.266 & 59.727 & 70.307 & 67.303 \\
Shin & 5623 & 63.103 & 60.862 & $+2.241$ & 61.716 & 59.873 & 69.889 & 66.621 \\
Shin & 6737 & 63.276 & 60.977 & $+2.299$ & 61.655 & 59.990 & 69.824 & 66.670 \\
\bottomrule
\end{tabular}
\end{table*}

Across the 25 target--seed cells, dynamic BA exceeded static BA in 22. All three reversals occurred in BNCI, where effects ranged from $-2.045$ to $+2.392$~pp; every seed was positive in Cho, Lee, PhysioNet, and Shin. The pooled SSM estimate is positive but heterogeneous, and inference remains anchored at the target level rather than the seed level.

Table~\ref{tab:subject-heterogeneity} reports the participant-level distribution after averaging BA over the five seeds. Dynamic--BIOT differences were positive for 75.2--90.7\% of participants in each target. Dynamic--static differences were more variable, from 33.3\% positive in BNCI to 73.1\% in Cho. These fractions describe within-target heterogeneity and are not independent replications across domains.

\begin{table*}[h]
\centering
\caption{Subject-level heterogeneity. Entries are median [Q1,Q3] BA or paired BA differences in percentage points; ``$>0$'' gives the descriptive proportion of participants with a positive paired difference.}
\label{tab:subject-heterogeneity}
\scriptsize
\setlength{\tabcolsep}{2.2pt}
\resizebox{\textwidth}{!}{%
\begin{tabular}{lrrrrrrrr}
\toprule
Target & $n$ & Dynamic BA & Static BA & BIOT BA & Dynamic--BIOT & $>0$ & Dynamic--static & $>0$ \\
\midrule
BNCI & 9 & 61.94 [56.32,73.89] & 64.72 [54.79,78.12] & 53.82 [51.39,57.99] & 7.29 [4.65,9.31] & 77.8\% & $-1.04$ [$-3.26$,1.53] & 33.3\% \\
Cho & 52 & 56.88 [52.49,62.38] & 55.08 [51.46,61.45] & 49.90 [47.70,51.38] & 7.00 [2.44,13.30] & 82.7\% & 1.25 [$-0.03$,2.74] & 73.1\% \\
Lee & 54 & 58.08 [53.12,66.50] & 56.27 [52.64,64.75] & 51.62 [50.31,53.30] & 6.20 [3.10,13.24] & 90.7\% & 0.78 [$-0.09$,1.88] & 72.2\% \\
PhysioNet & 109 & 56.25 [51.61,64.35] & 54.60 [50.00,62.83] & 51.28 [48.51,55.24] & 5.89 [0.30,12.87] & 75.2\% & 1.13 [$-1.07$,3.89] & 61.5\% \\
Shin & 29 & 61.33 [54.67,71.33] & 59.00 [53.33,69.67] & 51.67 [48.33,56.67] & 10.00 [1.00,18.00] & 79.3\% & 2.33 [0.00,4.67] & 72.4\% \\
\bottomrule
\end{tabular}}
\end{table*}

Omitting one target at a time left the dynamic--BIOT estimate between 7.618 and 8.600~pp and the dynamic--static estimate between 0.792 and 1.470~pp (Table~\ref{tab:lodo-sensitivity}). Each leave-one-target-out value is the arithmetic mean of the remaining four target effects. The primary interval is unchanged and comes from the dataset--subject hierarchical bootstrap over all five targets.

\begin{table*}[h]
\centering
\caption{Leave-one-target-out sensitivity, in percentage points. ``Full'' is the five-target equal mean with its primary hierarchical 95\% interval; the other columns omit the indicated target.}
\label{tab:lodo-sensitivity}
\scriptsize
\setlength{\tabcolsep}{4pt}
\begin{tabular}{lrrrrrr}
\toprule
Contrast & Full [95\% CI] & omit BNCI & omit Cho & omit Lee & omit PhysioNet & omit Shin \\
\midrule
Dynamic--BIOT & 8.182 [6.382,10.196] & 8.205 & 8.206 & 8.279 & 8.600 & 7.618 \\
Dynamic--static & 1.059 [0.013,1.892] & 1.470 & 0.985 & 1.106 & 0.941 & 0.792 \\
\bottomrule
\end{tabular}
\end{table*}

\FloatBarrier
\section{Post-Score-Seal Equal-Information FI/SSI Controls}
\label{app:equal-info}

\begin{figure*}[h]
  \centering
  \includegraphics[width=\textwidth]{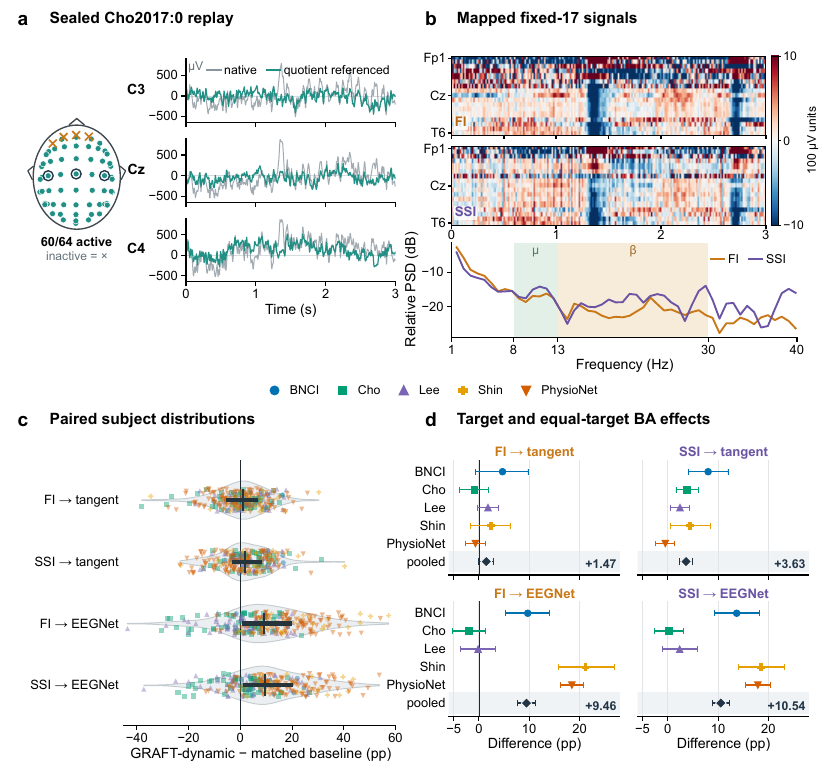}
  \caption{\textbf{Post-score-seal equal-information FI/SSI controls.} \textbf{a}, Active-sensor geometry and native versus quotient-referenced C3, Cz, and C4 traces from the sealed Cho2017:0 replay; 60 of 64 sensors were active, and all traces are shown on the same physical scale. \textbf{b}, Fixed-17 FI and SSI mappings for the same label-blind trial after active-native common-average referencing, mask-specific mapping, fixed-17 common-average referencing, scaling in 100-$\mu$V units, and clipping at $\pm 10$. The lower panel shows Welch relative power spectral density averaged over C3, Cz, and C4 from 1 to 40~Hz. \textbf{c}, Distributions of balanced-accuracy differences for all 253 paired subjects under each matched control. Each subject contributes once per control; colours and symbols denote target datasets, and black bars show the median and interquartile range. \textbf{d}, Target-level paired estimates with 95\% subject-bootstrap intervals and equal-target subject-cluster pooled effects from 20,000 bootstrap replicates. Positive values favour dynamic GRAFT-EEG. All controls covered 5/5 targets (253 unique subjects; 1,012 control-specific paired rows); tangent used one registered split seed and EEGNet used five seeds. These post-score-seal controls do not revise the primary inference.}
  \label{fig:equal-info}
\end{figure*}

All four matched point estimates favored dynamic \method: 1.47~pp for FI+tangent (95\% CI $-0.07$ to 2.96), 3.63~pp for SSI+tangent (2.32 to 4.95), 9.46~pp for FI+EEGNet (7.68 to 11.25), and 10.54~pp for SSI+EEGNet (8.90 to 12.16). FI+tangent crossed zero. The supplementary aggregate contains 1,012 paired subject rows across 20 target contrasts and 253 participants, with 20,000 target-stratified subject-cluster bootstrap replicates. All 60 fits, raw predictions, and score manifests were present; four group audits and the independent aggregate audit passed with $P0=P1=0$. Because this aggregate was defined after the primary scores were frozen, it is supplementary and does not revise the primary inference.

\begin{table*}[h]
\centering
\caption{Target-resolved equal-information controls. Values are subject-mean BA percentages. ``Effect'' is dynamic \method minus the matched control. Tangent controls retain their registered single-seed denominator; EEGNet controls retain five seeds.}
\label{tab:fi-ssi-target}
\scriptsize
\setlength{\tabcolsep}{3.5pt}
\begin{tabular}{llrrrr}
\toprule
Head & Target & Subjects & Matched control & \method dynamic & Effect (pp) \\
\midrule
FI + tangent & BNCI & 9 & 59.645 & 64.336 & $+4.691$ \\
 & Cho & 52 & 59.032 & 58.172 & $-0.860$ \\
 & Lee & 54 & 58.519 & 60.318 & $+1.799$ \\
 & PhysioNet & 109 & 58.930 & 58.247 & $-0.683$ \\
 & Shin & 29 & 60.862 & 63.253 & $+2.391$ \\
\midrule
SSI + tangent & BNCI & 9 & 56.327 & 64.336 & $+8.009$ \\
 & Cho & 52 & 54.379 & 58.172 & $+3.793$ \\
 & Lee & 54 & 57.894 & 60.318 & $+2.424$ \\
 & PhysioNet & 109 & 58.738 & 58.247 & $-0.491$ \\
 & Shin & 29 & 58.851 & 63.253 & $+4.402$ \\
\midrule
FI + EEGNet & BNCI & 9 & 54.637 & 64.336 & $+9.699$ \\
 & Cho & 52 & 60.085 & 58.172 & $-1.914$ \\
 & Lee & 54 & 60.464 & 60.318 & $-0.146$ \\
 & PhysioNet & 109 & 39.776 & 58.247 & $+18.471$ \\
 & Shin & 29 & 42.080 & 63.253 & $+21.172$ \\
\midrule
SSI + EEGNet & BNCI & 9 & 50.656 & 64.336 & $+13.681$ \\
 & Cho & 52 & 57.920 & 58.172 & $+0.251$ \\
 & Lee & 54 & 57.959 & 60.318 & $+2.358$ \\
 & PhysioNet & 109 & 40.362 & 58.247 & $+17.885$ \\
 & Shin & 29 & 44.747 & 63.253 & $+18.506$ \\
\bottomrule
\end{tabular}
\end{table*}

\begin{figure*}[!t]
  \centering
  \includegraphics[width=0.96\textwidth]{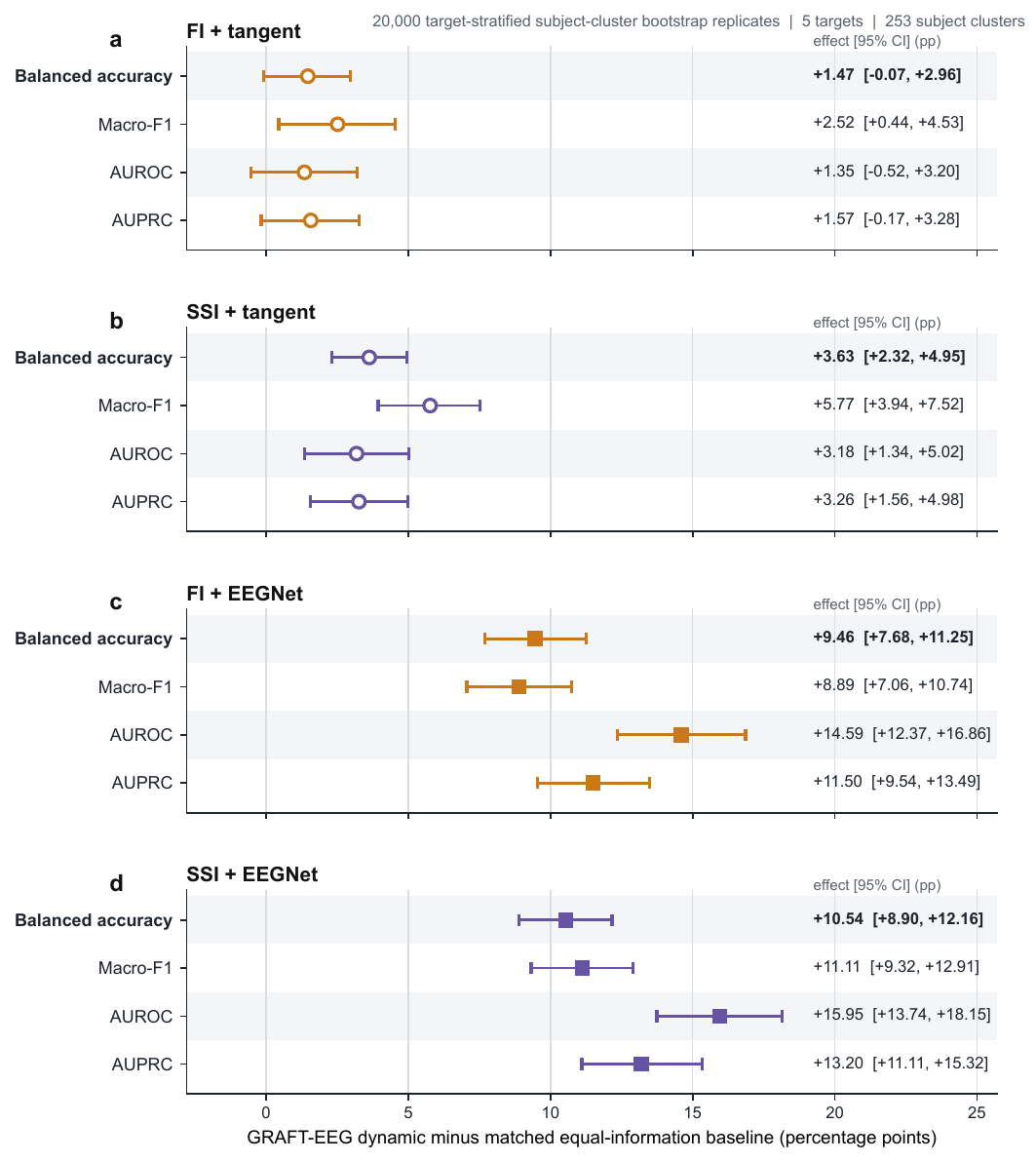}
  \caption{\textbf{All-metric equal-information inference.} Each panel reports dynamic \method minus one matched equal-information control for balanced accuracy, Macro-F1, AUROC, and AUPRC. Points are equal-target effects and bars are 95\% target-stratified subject-cluster bootstrap intervals (20,000 replicates; five targets; 253 subject clusters; audit seed 20260824). Balanced accuracy, shown in bold, was the prespecified primary metric for this supplementary analysis. Hollow circles denote tangent-space controls; filled squares denote EEGNet controls. The exact plotted estimates and intervals are printed at right and are taken directly from the sealed 16-row source table without refitting or resampling.}
  \label{fig:fi-ssi-all-metrics}
\end{figure*}

\FloatBarrier

\section{Calibration, Stress, and Execution Coverage}
\label{app:robustness}

\begin{figure*}[h]
  \centering
  \includegraphics[width=0.91\textwidth]{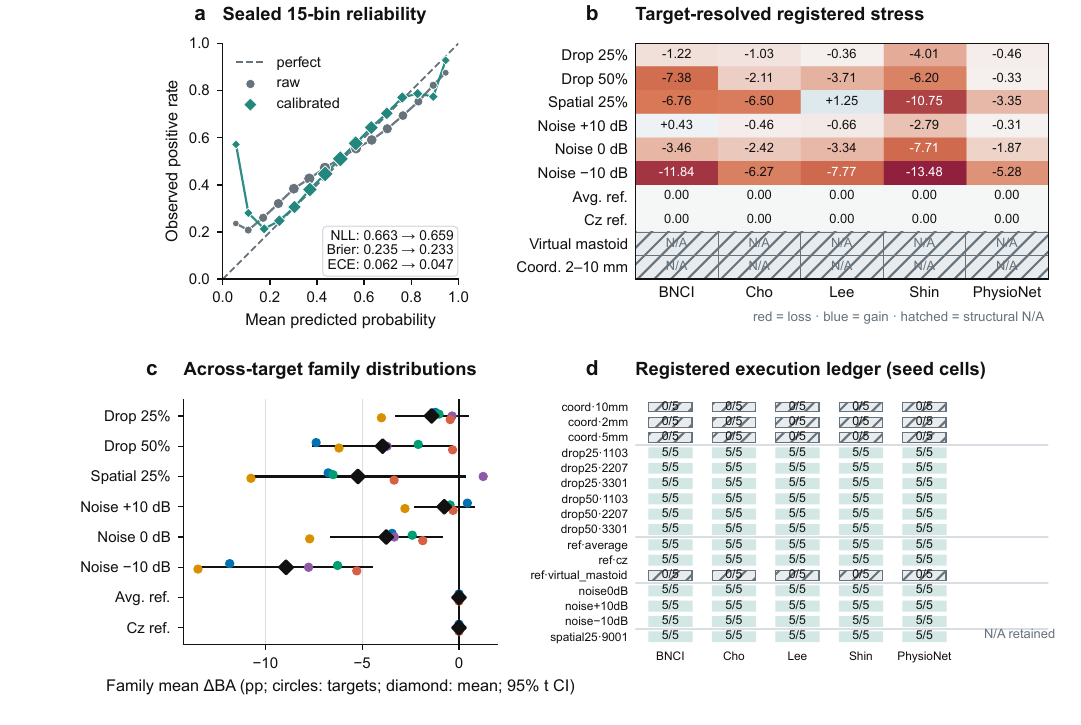}
  \caption{\textbf{Calibration and registered stress.} \textbf{a}, Trial-pooled 15-bin reliability curves reconstructed from all 25 sealed target--seed cells, together with equal-target NLL, Brier, and ECE. \textbf{b}, Target-resolved BA changes in percentage points under the registered stress families; hatched entries are structural N/A. \textbf{c}, The five target effects in percentage points within each family, with the equal-target mean and 95\% $t$ interval. \textbf{d}, The seed-cell execution ledger; structural N/A entries retain their zero numerator and are not converted into effects.}
  \label{fig:robustness}
\end{figure*}

Source-only calibration reduced equal-target ECE from 0.0617 to 0.0474, Brier score from 0.2346 to 0.2330, and negative log likelihood from 0.6631 to 0.6585, without changing BA, AUROC, or AUPRC. Mean BA changes were $-1.42$~pp for 25\% channel dropout, $-3.95$ for 50\% dropout, $-5.22$ for 25\% spatial dropout, and $-0.76$, $-3.76$, and $-8.93$~pp for +10, 0, and $-10$~dB noise. Virtual mastoid (0/25) and the registered coordinate row (0/75) were structurally unavailable in this historical E4--E6 namespace and remain N/A below. A later sealed extension reused 75 completed tangent cells from a separate legacy namespace and executed 175 additional coordinate cells (Appendix~\ref{app:coordinate-perturbation}); it does not overwrite this ledger. Cross-seed stability contrasts were all positive, indicating greater rather than lower seed variability for \method; this negative result is retained.

\begin{table*}[h]
\centering
\caption{Target-resolved source-only calibration. Lower is better for NLL, Brier, and ECE. Each entry averages five frozen target cells; calibration parameters were fitted on source predictions only.}
\label{tab:calibration-target}
\scriptsize
\setlength{\tabcolsep}{4pt}
\begin{tabular}{lrrrrrr}
\toprule
Target & NLL raw & NLL cal. & Brier raw & Brier cal. & ECE raw & ECE cal. \\
\midrule
BNCI & \cellcolor{GraftBlue!6}\textbf{0.6329} & \cellcolor{GraftOrange!10}0.6378 & \cellcolor{GraftBlue!6}\textbf{0.2212} & \cellcolor{GraftOrange!10}0.2233 & \cellcolor{GraftBlue!6}\textbf{0.0649} & \cellcolor{GraftOrange!10}0.0760 \\
Cho & 0.6883 & \cellcolor{GraftTeal!10}\textbf{0.6746} & 0.2457 & \cellcolor{GraftTeal!10}\textbf{0.2404} & 0.0741 & \cellcolor{GraftTeal!10}\textbf{0.0342} \\
Lee & 0.6597 & \cellcolor{GraftTeal!10}\textbf{0.6568} & 0.2338 & \cellcolor{GraftTeal!10}\textbf{0.2325} & 0.0459 & \cellcolor{GraftTeal!10}\textbf{0.0317} \\
PhysioNet & 0.6953 & \cellcolor{GraftTeal!10}\textbf{0.6754} & 0.2479 & \cellcolor{GraftTeal!10}\textbf{0.2408} & 0.0844 & \cellcolor{GraftTeal!10}\textbf{0.0374} \\
Shin & \cellcolor{GraftBlue!6}\textbf{0.6394} & \cellcolor{GraftOrange!10}0.6480 & \cellcolor{GraftBlue!6}\textbf{0.2243} & \cellcolor{GraftOrange!10}0.2281 & \cellcolor{GraftBlue!6}\textbf{0.0390} & \cellcolor{GraftOrange!10}0.0578 \\
\midrule
\rowcolor{GraftGray!7}
\textbf{Equal-target} & 0.6631 & \cellcolor{GraftTeal!12}\textbf{0.6585} & 0.2346 & \cellcolor{GraftTeal!12}\textbf{0.2330} & 0.0617 & \cellcolor{GraftTeal!12}\textbf{0.0474} \\
\bottomrule
\end{tabular}
\end{table*}

Calibration improves all three pooled probability-error summaries, but not every target: BNCI and Shin worsen on all three. Discrimination is unchanged by a monotone temperature map: equal-target BA remains 60.865\%, AUROC 66.447\%, and AUPRC 67.531\% before and after calibration. These unchanged values are retained in the source table rather than repeated in every target row.

\begin{table*}[h]
\centering
\caption{Target-resolved robustness family matrix. Values are dynamic BA change in percentage points relative to the unperturbed target cell. ``Exec.'' is complete/registered seed cells. Structural N/A is not coded as zero.}
\label{tab:robustness-matrix}
\scriptsize
\setlength{\tabcolsep}{3.2pt}
\begin{tabular}{lrrrrrrc}
\toprule
Condition family & BNCI & Cho & Lee & PhysioNet & Shin & Equal-target & Exec. \\
\midrule
Random channel drop 25\% & $-1.22$ & $-1.03$ & $-0.36$ & $-0.46$ & $-4.01$ & $-1.42$ & 75/75 \\
Random channel drop 50\% & $-7.38$ & $-2.11$ & $-3.71$ & $-0.33$ & $-6.20$ & $-3.95$ & 75/75 \\
Spatial contiguous drop 25\% & $-6.76$ & $-6.50$ & $+1.25$ & $-3.35$ & $-10.75$ & $-5.22$ & 25/25 \\
Noise +10 dB & $+0.43$ & $-0.46$ & $-0.66$ & $-0.31$ & $-2.79$ & $-0.76$ & 25/25 \\
Noise 0 dB & $-3.46$ & $-2.42$ & $-3.34$ & $-1.87$ & $-7.71$ & $-3.76$ & 25/25 \\
Noise $-10$ dB & $-11.84$ & $-6.27$ & $-7.77$ & $-5.28$ & $-13.48$ & $-8.93$ & 25/25 \\
Average reference & $0.00$ & $0.00$ & $0.00$ & $0.00$ & $0.00$ & $0.00$ & 25/25 \\
Cz reference & $0.00$ & $0.00$ & $0.00$ & $0.00$ & $0.00$ & $0.00$ & 25/25 \\
\rowcolor{GraftGray!12}
Virtual mastoid & \textbf{N/A} & \textbf{N/A} & \textbf{N/A} & \textbf{N/A} & \textbf{N/A} & \textbf{N/A} & \textbf{0/25} \\
\rowcolor{GraftGray!12}
Coordinate shift 2--10 mm & \textbf{N/A} & \textbf{N/A} & \textbf{N/A} & \textbf{N/A} & \textbf{N/A} & \textbf{N/A} & \textbf{0/75} \\
\bottomrule
\end{tabular}
\end{table*}

\begin{figure*}[h]
  \centering
  \includegraphics[width=0.94\textwidth]{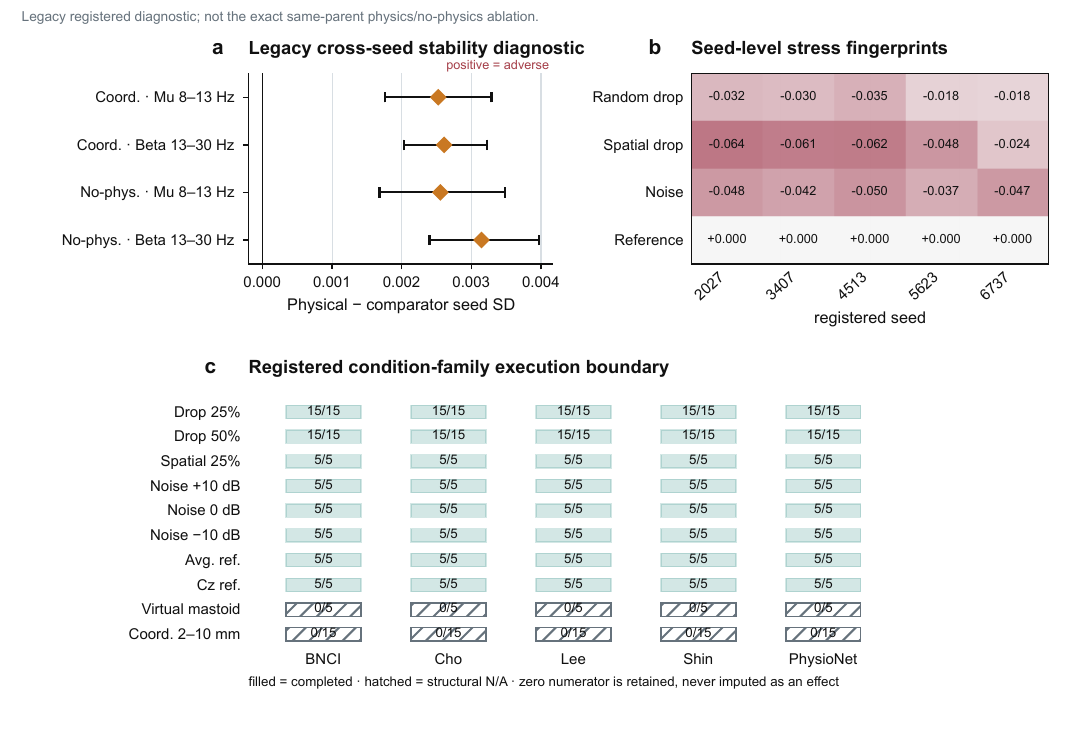}
  \caption{\textbf{Legacy stability and execution boundaries.} This registered diagnostic is separate from the exact same-parent physical-leadfield control in Fig.~\ref{fig:main-results}. \textbf{a}, Physical-minus-comparator cross-seed standard-deviation contrasts for coordinate and legacy no-physics diagnostics. All four estimates are positive; negative would mean that the physical model is more stable, so the observed direction is adverse rather than supportive evidence. \textbf{b}, Seed-level mean BA-change fingerprints for random dropout, spatial dropout, noise, and reference families. \textbf{c}, Completed/registered cells by target and condition family. Structural N/A remains hatched with its zero numerator and is never imputed as an effect.}
  \label{fig:stability-execution}
\end{figure*}

\FloatBarrier
\section{Few-Shot Target-Label Adaptation Boundary}
\label{app:fewshot}

The few-shot analysis relaxes the original access contract by revealing 1\%, 5\%, or 10\% of target labels only after the zero-target model has been frozen. Four adaptation strategies are then compared on disjoint evaluation trials. These results define an adaptation boundary and should not be read as target-statistics-blind performance. Each strategy--fraction pair completed all 25 target--seed cells (300 rows in total), and the independent audit passed.

\begin{table*}[h]
\centering
\caption{Target-resolved few-shot BA (\%). Label fractions denote target trials used for adaptation; all remaining trials form a disjoint evaluation set.}
\label{tab:fewshot-target}
\scriptsize
\setlength{\tabcolsep}{3.5pt}
\begin{tabular}{lcrrrrrr}
\toprule
Strategy & Labels & BNCI & Cho & Lee & PhysioNet & Shin & Equal-target \\
\midrule
Frozen linear probe & 1\% & 58.939 & 52.879 & 56.608 & 53.211 & 54.060 & 55.139 \\
 & 5\% & 61.231 & 56.606 & 59.947 & 53.949 & 57.075 & 57.762 \\
 & 10\% & 63.718 & 57.856 & 60.804 & 54.934 & 58.774 & 59.217 \\
\midrule
Full GRAFT head fine-tune & 1\% & 66.102 & 58.309 & 60.736 & 57.776 & 63.692 & 61.323 \\
 & 5\% & 65.917 & 58.424 & 61.163 & 57.915 & 63.964 & 61.477 \\
 & 10\% & 66.887 & 58.821 & 61.398 & 57.888 & 63.614 & 61.722 \\
\midrule
Low-rank latent residual & 1\% & 61.316 & 54.921 & 57.227 & 54.265 & 57.162 & 56.978 \\
 & 5\% & 64.654 & 55.525 & 59.227 & 53.738 & 59.253 & 58.479 \\
 & 10\% & 63.777 & 56.182 & 59.784 & 53.328 & 58.506 & 58.315 \\
\midrule
PEFT dynamic readout & 1\% & 65.761 & 58.027 & 60.678 & 57.969 & 63.215 & 61.130 \\
 & 5\% & 65.411 & 57.939 & 60.729 & 57.880 & 63.570 & 61.106 \\
 & 10\% & 66.213 & 58.048 & 60.674 & 57.663 & 63.474 & 61.214 \\
\bottomrule
\end{tabular}
\end{table*}

\begin{table*}[h]
\centering
\caption{Equal-target few-shot metrics (\%). ECE is the 15-bin expected calibration error for the positive class. Each row summarizes 25 completed target--seed cells.}
\label{tab:fewshot-pooled}
\scriptsize
\setlength{\tabcolsep}{4pt}
\begin{tabular}{lcrrrrc}
\toprule
Strategy & Labels & BA & Macro-F1 & AUROC & ECE & Cells \\
\midrule
Frozen linear probe & 1\% & 55.139 & 54.872 & 57.149 & 28.017 & 25/25 \\
 & 5\% & 57.762 & 57.713 & 60.836 & 19.790 & 25/25 \\
 & 10\% & 59.217 & 59.202 & 63.131 & 13.730 & 25/25 \\
\midrule
Full GRAFT head fine-tune & 1\% & 61.323 & 61.189 & 66.096 & 4.772 & 25/25 \\
 & 5\% & 61.477 & 61.379 & 66.221 & 3.898 & 25/25 \\
 & 10\% & 61.722 & 61.701 & 66.594 & 3.477 & 25/25 \\
\midrule
Low-rank latent residual & 1\% & 56.978 & 56.815 & 61.401 & 13.259 & 25/25 \\
 & 5\% & 58.479 & 58.400 & 62.466 & 12.143 & 25/25 \\
 & 10\% & 58.315 & 58.268 & 62.566 & 13.784 & 25/25 \\
\midrule
PEFT dynamic readout & 1\% & 61.130 & 60.973 & 65.863 & 4.732 & 25/25 \\
 & 5\% & 61.106 & 60.949 & 65.752 & 4.522 & 25/25 \\
 & 10\% & 61.214 & 61.173 & 65.798 & 4.205 & 25/25 \\
\bottomrule
\end{tabular}
\end{table*}

The small separation between full-head and PEFT adaptation is not used to revise the zero-target model. At 1\% labels their equal-target BA values are 61.323\% and 61.130\%, while the frozen linear probe is 55.139\%. Increasing the frozen-probe fraction improves its BA monotonically, whereas the low-rank residual is not monotone between 5\% and 10\%; both patterns are retained without a post hoc winner claim.

\FloatBarrier

\section{Separately Sealed External Evaluation}
\label{app:external}

\begin{figure*}[!t]
  \centering
  \includegraphics[width=0.98\textwidth]{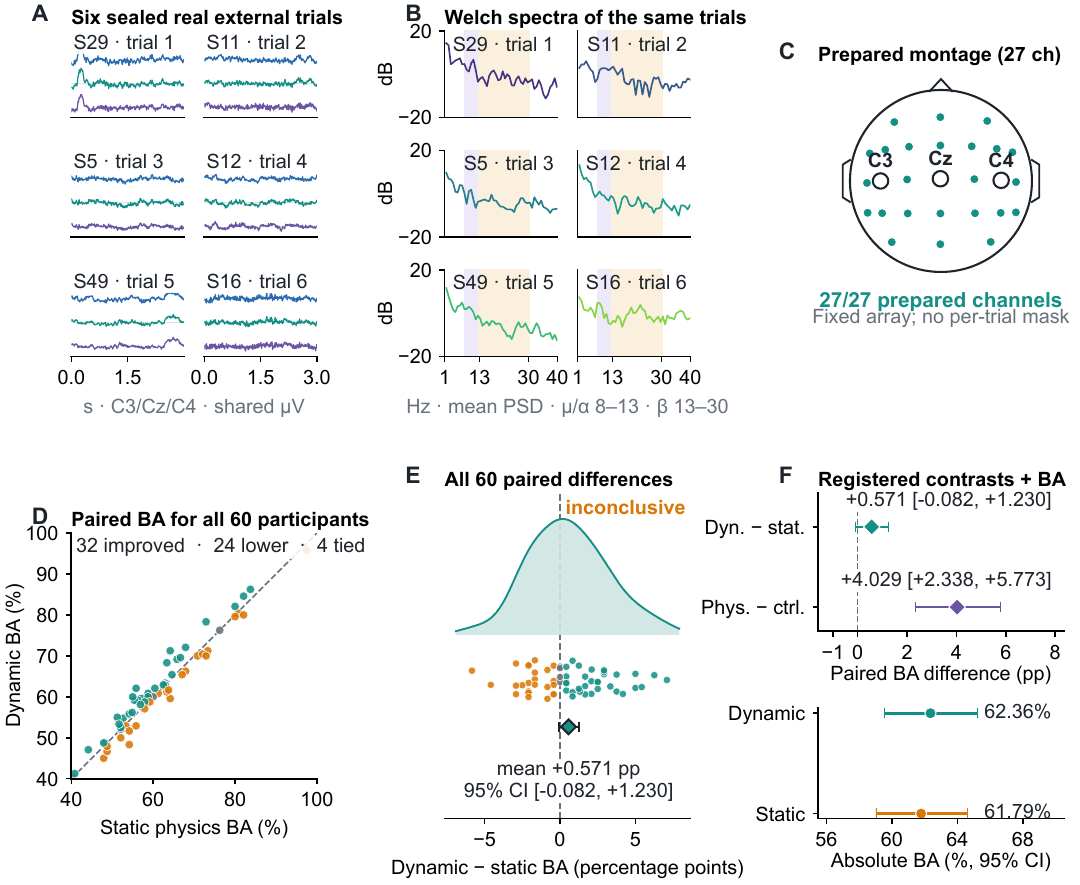}
  \caption{\textbf{One-shot external evaluation on 60 held-out participants.} \textbf{a}, C3/Cz/C4 waveforms from six label- and prediction-blind SHA-selected real Dreyer2023A trials. \textbf{b}, Welch spectra of the same six trials. \textbf{c}, The prepared 27-channel montage; the prepared export has no per-trial mask. \textbf{d}, Paired dynamic and static BA for all 60 participants. \textbf{e}, The complete participant-level dynamic--static distribution with the sealed mean and interval. \textbf{f}, Two preregistered contrasts and descriptive absolute BA estimates. The dynamic--static contrast is inconclusive; the physical-operator contrast is positive.}
  \label{fig:external}
\end{figure*}

Dreyer2023A is supported by a protocol seal, score seal, replay audit, and independent final audit. Of 14,312 prepared trials, 14,309 passed pre-lock signal QC and entered one-time scoring. Dynamic BA was 62.36\% (59.56--65.24) and static BA was 61.79\% (59.03--64.64). The dynamic--static contrast was $+0.57$~pp ($-0.08$ to $+1.23$; inconclusive), whereas physics--no-physics was $+4.03$~pp ($+2.34$ to $+5.77$; positive). The current coverage registration links these artifacts with $P0=P1=0$. Dreyer therefore contributes an available one-shot external evaluation with \emph{partial-confirmation} status, but it does not qualify as unrestricted independent external validation.

\begin{table*}[!t]
\centering
\caption{Dreyer2023A one-shot external summary. Absolute BA rows are descriptive; the two contrast rows report the preregistered decisions. Values are percentages or percentage points with 95\% intervals.}
\label{tab:external-summary}
\small
\setlength{\tabcolsep}{5pt}
\begin{tabular}{llrrrl}
\toprule
Result type & Quantity & Estimate & Lower & Upper & Decision \\
\midrule
Absolute BA & \method dynamic & 62.359 & 59.562 & 65.239 & descriptive \\
Absolute BA & Static physics & 61.789 & 59.033 & 64.638 & descriptive \\
Preregistered contrast & Dynamic $-$ static & $+0.571$ & $-0.082$ & $+1.230$ & inconclusive \\
Preregistered contrast & Full physics $-$ no physics & $+4.029$ & $+2.338$ & $+5.773$ & positive \\
\bottomrule
\end{tabular}
\end{table*}

The registration manifest records the coverage ledger together with the external-registration, module, and independent-audit hashes, and sets \texttt{dreyer\_is\_unrestricted\_external\_validation=false}. This machine-readable constraint matches the partial-confirmation interpretation reported in Table~\ref{tab:external-summary}.

\FloatBarrier

\section{Known-source Validation and Condition Ledger}
\label{app:known-source}

\begin{figure*}[h]
  \centering
  \includegraphics[width=0.98\textwidth]{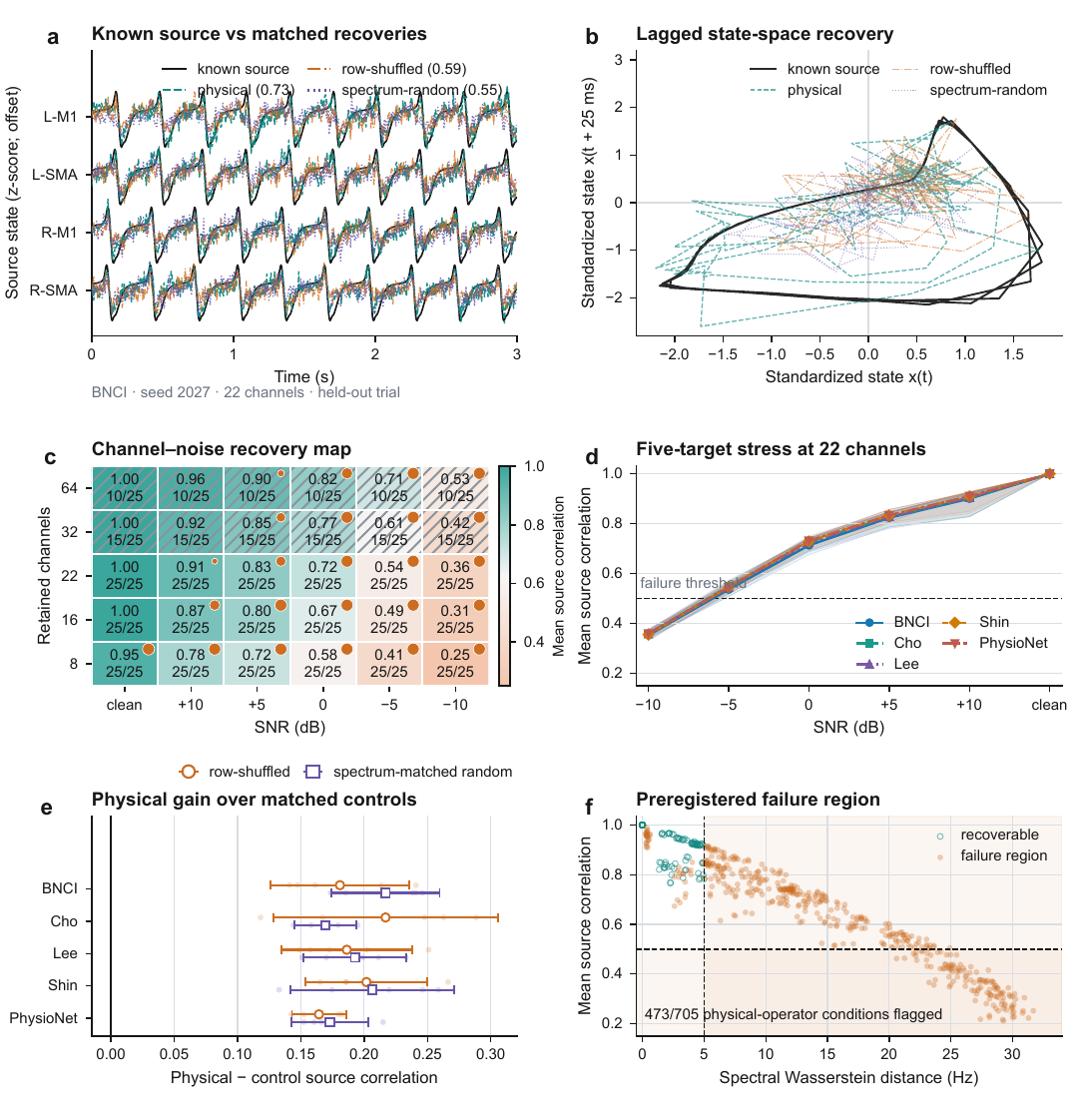}
  \caption{\textbf{Jansen--Rit validation with known sources.} \textbf{a}, Locked source trajectories and matched physical, row-shuffled, and spectrum-matched recoveries for one held-out BNCI seed-2027 trial at a 22-channel, 0-dB condition. \textbf{b}, A 25-ms lag embedding of the same left-M1 trajectories. \textbf{c}, Mean source correlation across registered retained-channel and SNR conditions; each cell reports the mean and completed/25 count, hatching retains structural N/A, and orange area encodes the fraction entering the preregistered scientific failure region. \textbf{d}, Five-target stress trajectories at a 22-channel budget across the five sealed seeds. \textbf{e}, Physical-operator advantage over row-shuffled and spectrum-matched controls by target; intervals are descriptive 20,000-draw bootstrap intervals. \textbf{f}, Failure-region placement for the 705 executed physical-operator conditions: 600 channel/noise conditions, 25 physical negative controls, and 80 reference conditions; 473/705 entered the preregistered failure region. The full executed ledger additionally contains 25 row-shuffled and 25 spectrum-matched-random negative controls, all in the failure region, giving 523/755 overall. All trajectories and estimates come from the sealed known-source archive. The experiment evaluates the explicit inverse on synthetic known sources and does not imply that the frozen real-EEG classifier recovered cortical sources or supports clinical localization.}
  \label{fig:synthetic}
\end{figure*}

The condition ledger contains 925 entries: 755 were executable and 170 were structurally unavailable. Among executed conditions, 232 met the prespecified recovery criterion and 523 entered the scientific failure region. Across 25 target--seed pairs, the physical operator increased mean source correlation by 0.190 versus shuffled-row controls (95\% descriptive bootstrap interval, 0.174--0.206) and by 0.192 versus spectrum-matched random controls (0.179--0.204). These intervals describe figure-level uncertainty rather than additional confirmatory tests. The independent check returned \texttt{PASS} ($P0=P1=0$).

\begin{table*}[h]
\centering
\caption{Target-by-block accounting for the 925 Jansen--Rit conditions. ``Recoverable'' and ``failure'' partition the 755 executed conditions; structural N/A remains in the registered denominator.}
\label{tab:e7-target-ledger}
\scriptsize
\setlength{\tabcolsep}{4pt}
\begin{tabular}{llrrrrr}
\toprule
Target & Block & Registered & Executed & Structural N/A & Recoverable & Failure \\
\midrule
BNCI & Channel/noise & 150 & 90 & 60 & 15 & 75 \\
 & Negative control & 15 & 15 & 0 & 0 & 15 \\
 & Reference & 20 & 15 & 5 & 15 & 0 \\
\midrule
Cho & Channel/noise & 150 & 150 & 0 & 45 & 105 \\
 & Negative control & 15 & 15 & 0 & 0 & 15 \\
 & Reference & 20 & 15 & 5 & 15 & 0 \\
\midrule
Lee & Channel/noise & 150 & 120 & 30 & 30 & 90 \\
 & Negative control & 15 & 15 & 0 & 0 & 15 \\
 & Reference & 20 & 20 & 0 & 20 & 0 \\
\midrule
PhysioNet & Channel/noise & 150 & 150 & 0 & 45 & 105 \\
 & Negative control & 15 & 15 & 0 & 0 & 15 \\
 & Reference & 20 & 15 & 5 & 15 & 0 \\
\midrule
Shin & Channel/noise & 150 & 90 & 60 & 17 & 73 \\
 & Negative control & 15 & 15 & 0 & 0 & 15 \\
 & Reference & 20 & 15 & 5 & 15 & 0 \\
\bottomrule
\end{tabular}
\end{table*}

\begin{table*}[h]
\centering
\caption{Known-source execution totals. The complete channel-budget-by-SNR grid is shown in Figure~\ref{fig:synthetic}c and retained without rounding in the corresponding source-data table. ``Recoverable'' and ``failure'' partition executed conditions; structural N/A remains in the registered denominator.}
\label{tab:e7-block-ledger}
\small
\setlength{\tabcolsep}{8pt}
\begin{tabular}{lrrrrr}
\toprule
Block & Registered & Executed & N/A & Recoverable & Failure \\
\midrule
Channel/noise & 750 & 600 & 150 & 152 & 448 \\
Negative control & 75 & 75 & 0 & 0 & 75 \\
Reference & 100 & 80 & 20 & 80 & 0 \\
\midrule
Total & 925 & 755 & 170 & 232 & 523 \\
\bottomrule
\end{tabular}
\end{table*}

The negative-control block is kept outside the channel/noise operating region because failure is expected by construction. Structural N/A instead records a channel budget or reference construction that is unavailable for a given montage; it is not a numerical failure. Table~\ref{tab:e7-block-ledger} therefore separates registered, executed, recoverable, failed, and unavailable conditions.

\FloatBarrier

\section{Frozen Trial Replay and Source-Fit Parameter Anatomy}
\label{app:mechanistic-visuals}

The following two figures expose the executed computation without selecting a favorable prediction. Cho2017 was fixed because its 64-channel montage makes the active-set operation visible, and trial \texttt{Cho2017:0} was then selected as the lowest retained queue index with at least one inactive channel. Labels, prediction correctness, and performance were unavailable to this rule. The parameter view comes from one prespecified source-fit fold and does not use target EEG or target-derived statistics.

\begin{figure*}[!t]
  \centering
  \includegraphics[width=0.98\textwidth]{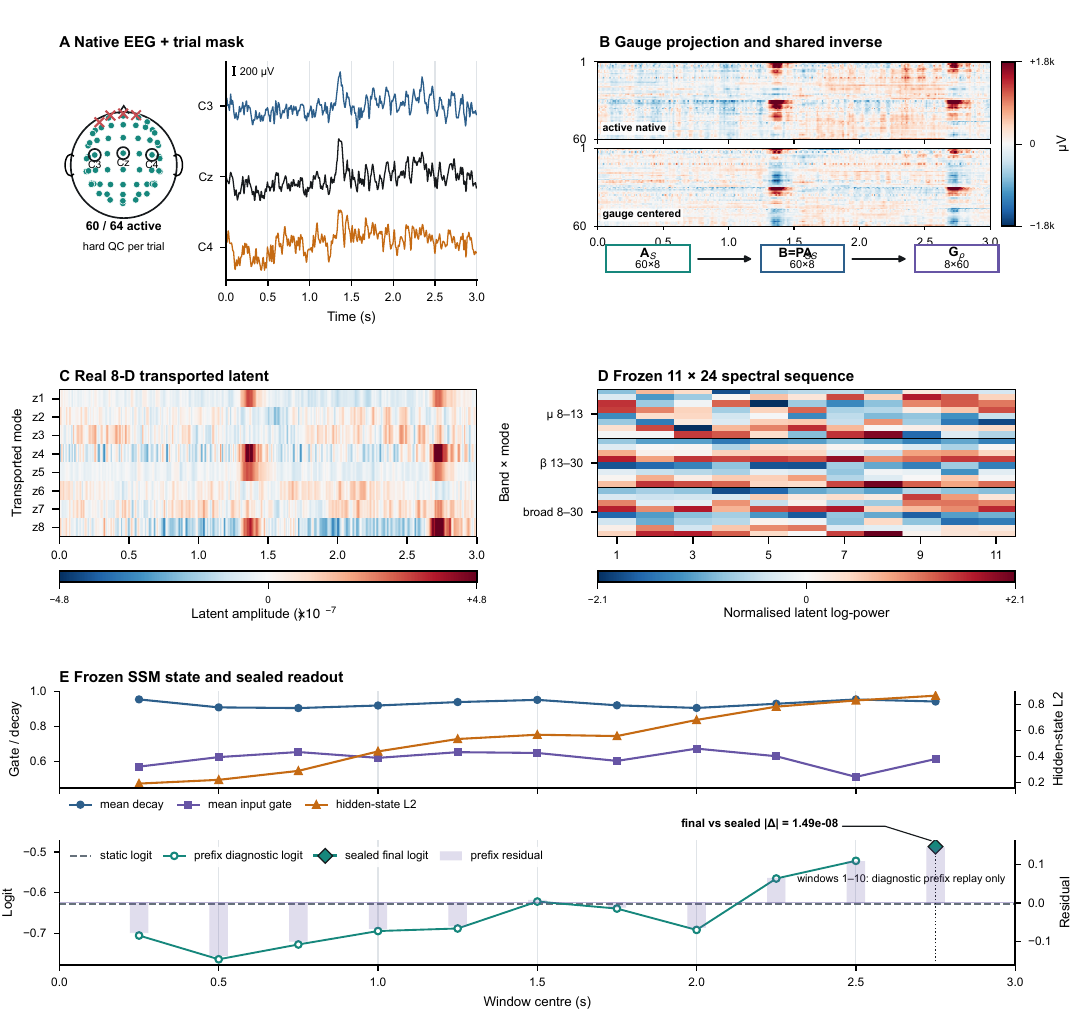}
  \caption{\textbf{Label-blind replay of a real retained trial.} \textbf{a}, Native 200-Hz Cho2017 trial \texttt{Cho2017:0}, its 64-channel geometry, trial-level hard-QC mask, and prespecified C3/Cz/C4 traces; 60 channels remain active. \textbf{b}, The same active signals before and after gauge projection, followed by the frozen active operator $A_S$, $B=P_SA_S$, and $G_\rho$. \textbf{c}, The resulting real $8\times600$ transported latent. \textbf{d}, The exact $11\times24$ windowed spectral sequence supplied to the frozen SSM. \textbf{e}, Mean decay, input gate, hidden-state norm, and a read-only prefix replay of the bounded residual and logit. Prefixes 1--10 are diagnostic only; the final point is the sole canonical scored output and matches the sealed logit within $1.49\times10^{-8}$. No label or outcome was used to choose the dataset, trial, channels, time range, axes, or colour limits.}
  \label{fig:real-trial}
\end{figure*}

\begin{figure*}[!t]
  \centering
  \includegraphics[width=0.96\textwidth]{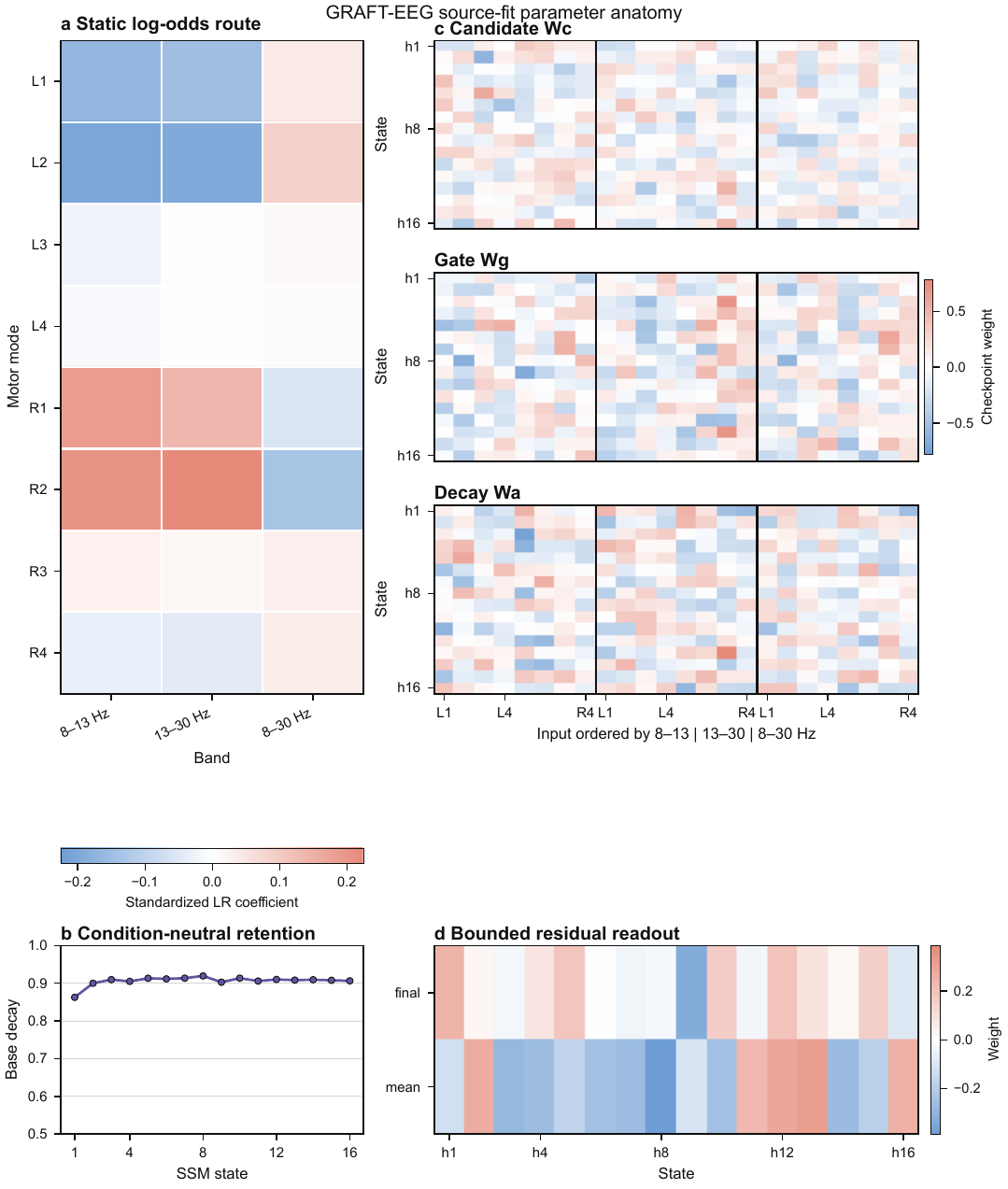}
  \caption{\textbf{Source-fitted parameter anatomy for one prespecified held-out-target fold.} \textbf{a}, Standardized logistic-regression coefficients across eight motor modes and three spectral bands. \textbf{b}, Condition-neutral state retention implied by the 16 frozen base-decay parameters. \textbf{c}, Frozen candidate, gate, and input-conditioned decay weights, with inputs ordered by band and motor mode. \textbf{d}, Frozen weights on terminal and sequence-mean states in the bounded residual readout. These panels describe stored source-fit parameters; they are not target-derived attribution maps or performance estimates.}
  \label{fig:parameter-anatomy}
\end{figure*}

\FloatBarrier

\section{Three-Dimensional Geometry-Conditioned Inverse Kernels}
\label{app:inverse-kernels}

\begin{figure*}[!t]
  \centering
  \includegraphics[width=0.96\textwidth]{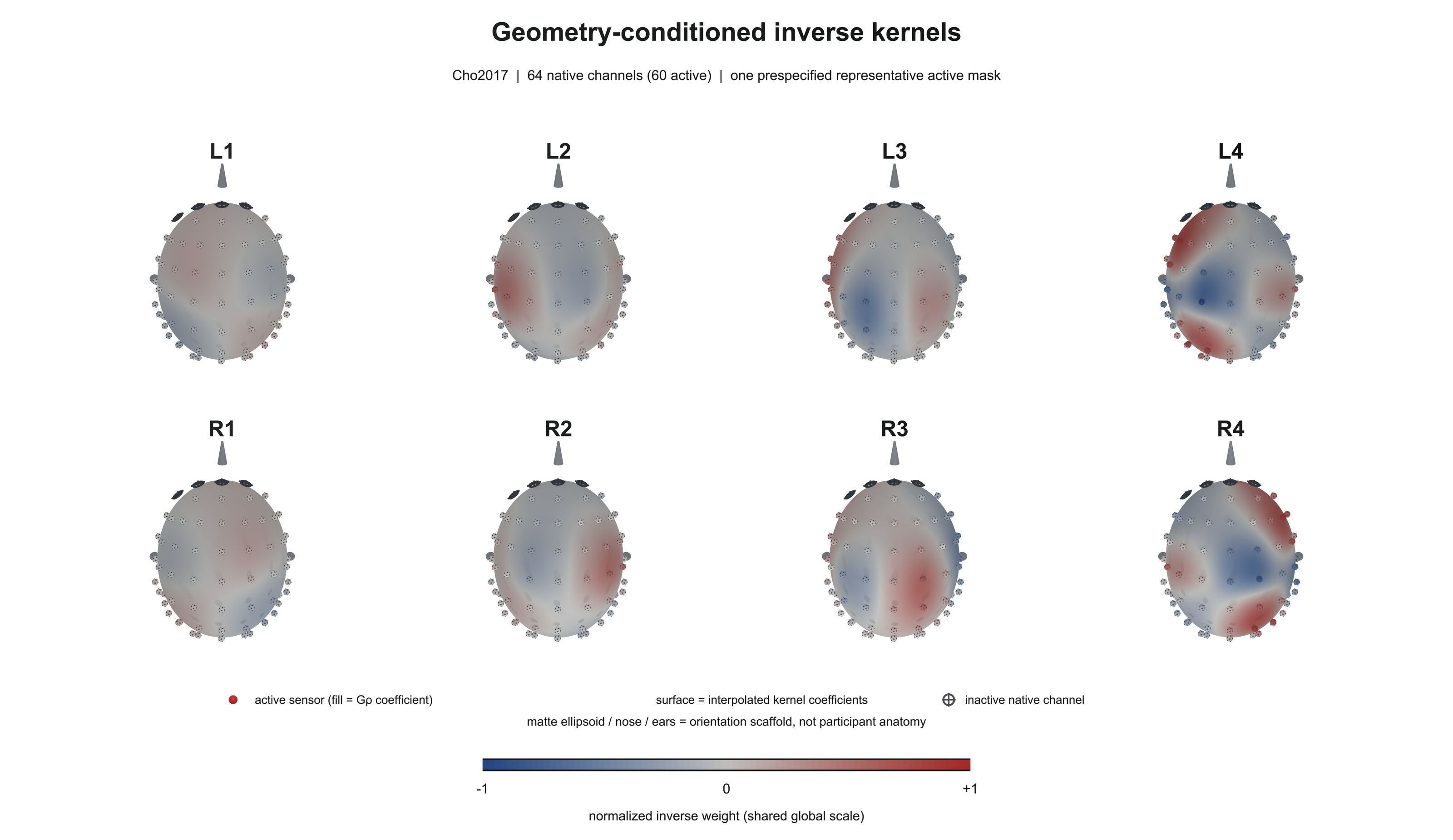}
  \caption{\textbf{Three-dimensional geometry-conditioned inverse kernels.} The eight panels show rows L1--L4 and R1--R4 of the locked Cho2017 inverse operator $G_\rho$ for one prespecified active mask (60 of 64 native channels). Sensor positions follow the recorded head-frame montage. Active sensors are colored by their corresponding $G_\rho$ coefficients, and the scalp surface interpolates the same values using one max-absolute normalization over the complete $8\times60$ matrix. Ring-and-cross markers identify the four inactive channels. The ellipsoid, nose, ears, lighting, and shadows provide orientation only and are not participant anatomy. Colors represent inverse-kernel coefficients---not learned Spat-Conv weights, cortical activity, source localization, statistical significance, or participant-specific anatomy.}
  \label{fig:inverse-kernels}
\end{figure*}

Figure~S9 uses eight scalp surfaces, 480 active-sensor markers, and 32 inactive-channel marker groups under one shared display scale. A companion source-data table lists every displayed coordinate, active-mask value, and raw and normalized $G_\rho$ coefficient. Surface interpolation is used only for rendering and does not enter training or evaluation.

\FloatBarrier
\section{Deterministically Selected Real-EEG Atlas}
\label{app:real-eeg-atlas}

The atlas exposes three real post-QC trials from each held-out dataset without selecting on labels, predictions, correctness, or performance. Distinct subjects are SHA-256 ranked within each dataset, followed by a SHA-256 ranking of eligible trials within subject. Each row binds the exact native montage and active mask to three central-motor traces, the full native channel--time array, and the mean active-channel Welch spectrum. Internal-dataset amplitudes are the executed fixed 100-$\mu$V model-input scale after common-average reference and trial de-meaning; they are not raw microvolts. Heatmap limits use one prespecified symmetric 99th-absolute-percentile rule per dataset across the three selected trials, while exact values remain in source data.

\begin{figure*}[!t]
  \centering
  \includegraphics[width=\textwidth]{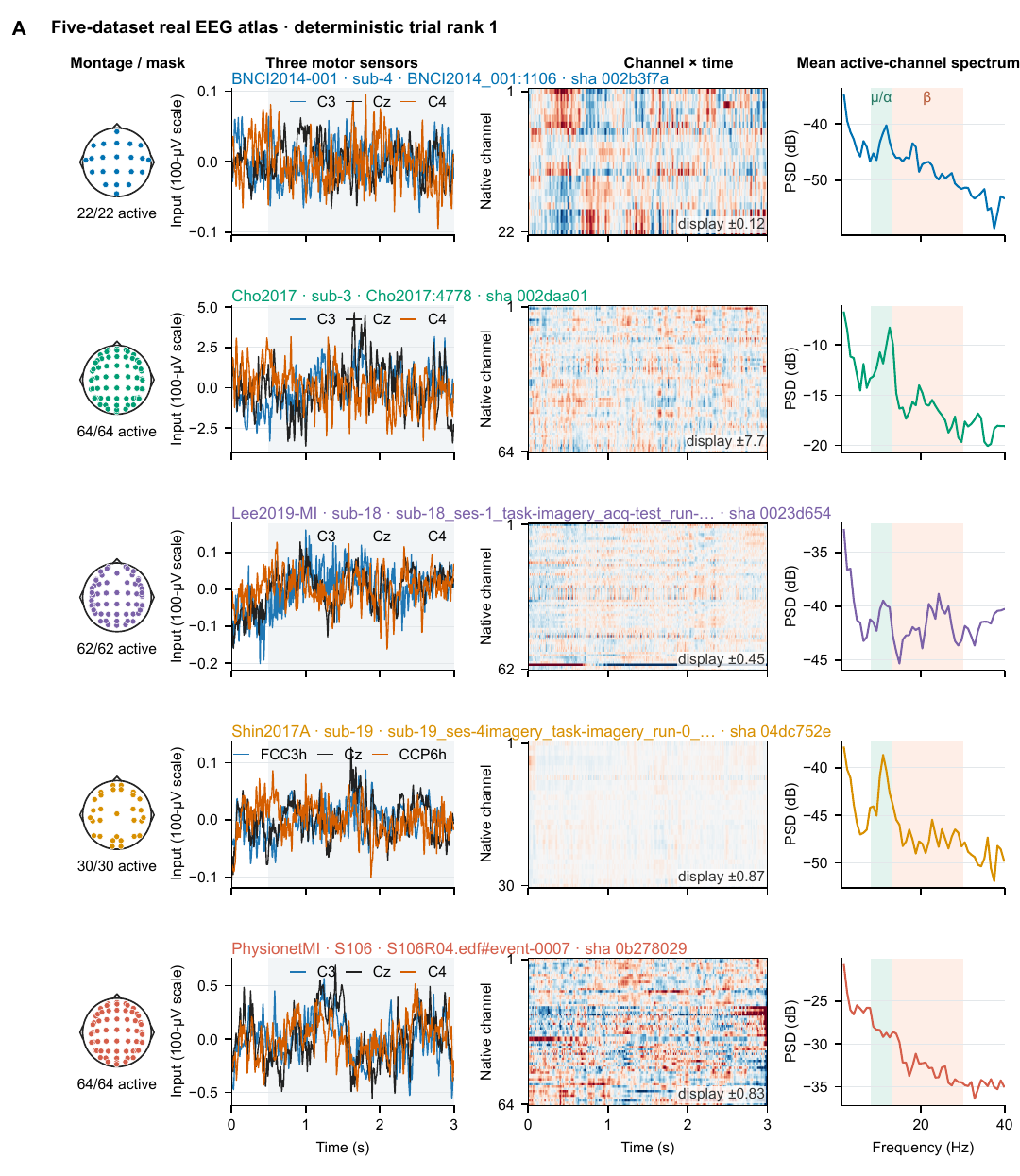}
  \caption{\textbf{Real-EEG atlas, deterministic rank 1.} One SHA-selected trial from each of BNCI2014-001, Cho2017, Lee2019-MI, Shin2017A, and PhysionetMI. Every dataset row shows the recorded montage and active set, three available central-motor sensors, the complete native-channel input, and the spectrum of the same trial. Shaded bands mark 8--13 and 13--30~Hz; no sleep-stage, event, ERD/ERS, or source-localization label is inferred.}
  \label{fig:real-eeg-atlas-1}
\end{figure*}

\begin{figure*}[!t]
  \centering
  \includegraphics[width=\textwidth]{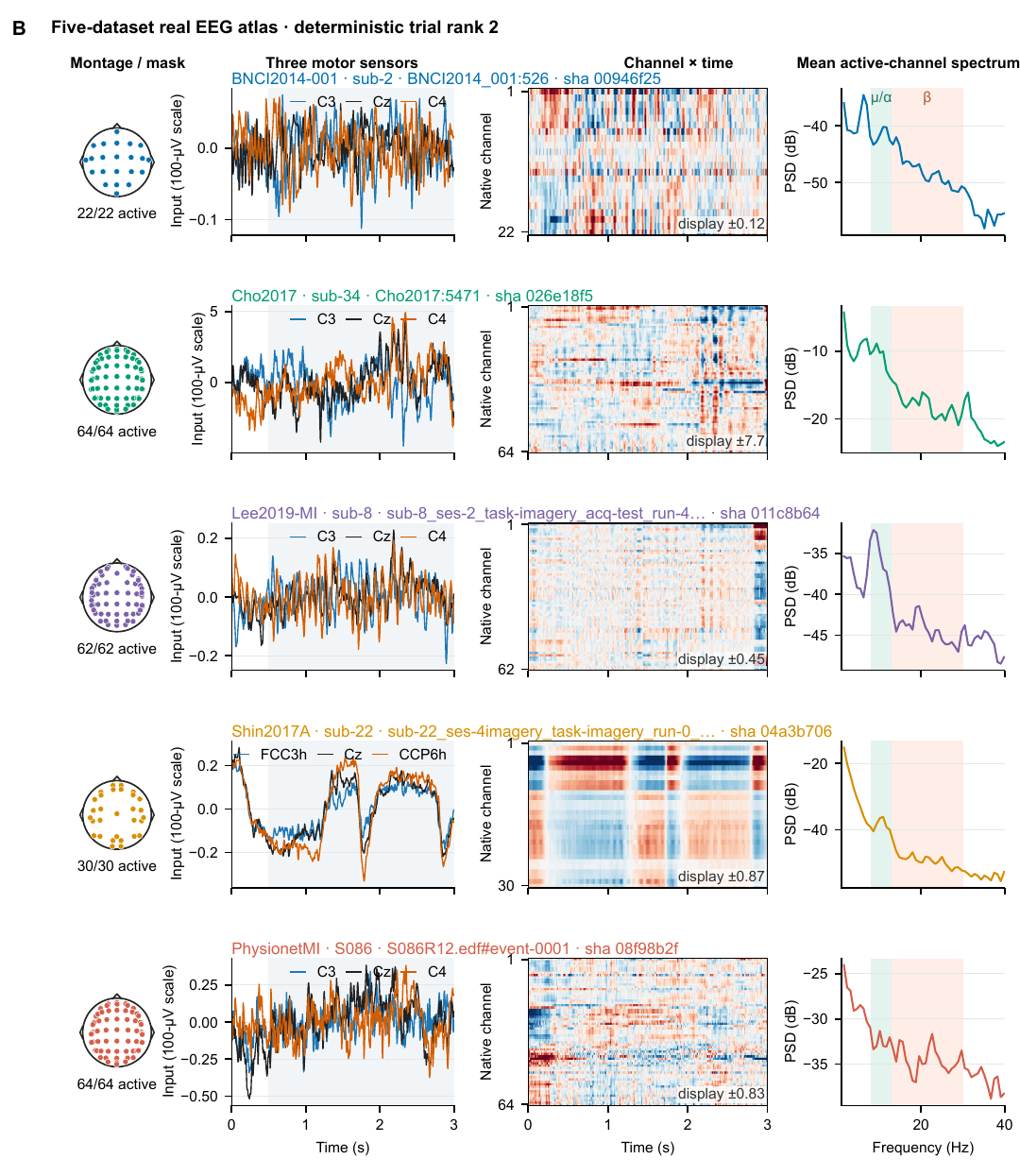}
  \caption{\textbf{Real-EEG atlas, deterministic rank 2.} The second independently SHA-ranked subject/trial for each dataset, rendered under the same locked channel, time, spectrum, colour-limit, and annotation rules as Figure~\ref{fig:real-eeg-atlas-1}. Trial identifiers and selection hashes are printed in the figure and retained in the source manifest.}
  \label{fig:real-eeg-atlas-2}
\end{figure*}

\begin{figure*}[!t]
  \centering
  \includegraphics[width=\textwidth]{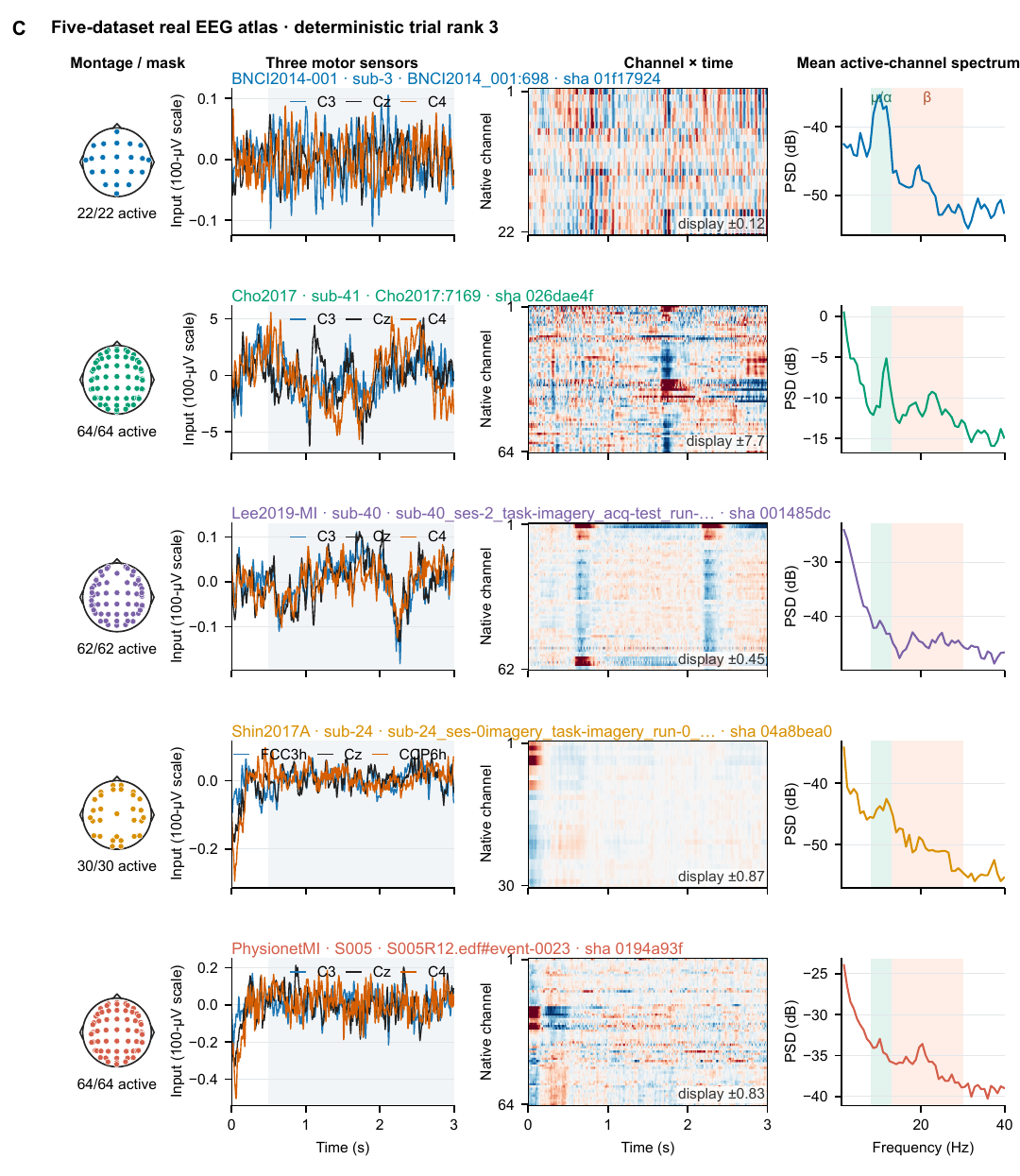}
  \caption{\textbf{Real-EEG atlas, deterministic rank 3.} The third SHA-ranked subject/trial for each dataset. Together, Figures~\ref{fig:real-eeg-atlas-1}--\ref{fig:real-eeg-atlas-3} show 15 real trials with complete trial IDs, native channel counts, selected sensor names, display limits, waveforms, and Welch source data.}
  \label{fig:real-eeg-atlas-3}
\end{figure*}

\FloatBarrier
\section{Strict Benchmark Matrix and Baseline Health}
\label{app:benchmark-matrix}

\begin{figure*}[!t]
  \centering
  \includegraphics[width=0.98\textwidth]{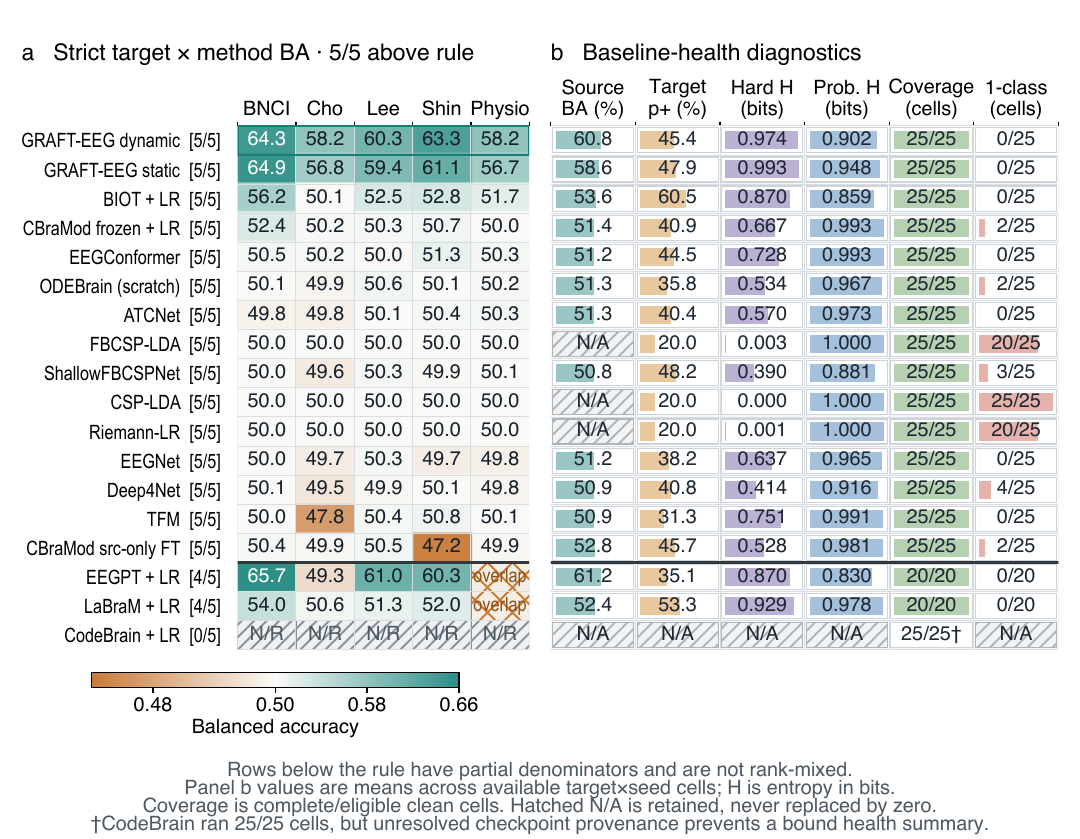}
  \caption{\textbf{Strict target--method matrix and baseline-health diagnostics.} \textbf{a}, Target-resolved BA for the 15 methods with a clean 5/5 denominator, followed below the rule by EEGPT and LaBraM with four clean targets and CodeBrain with unresolved checkpoint provenance. CodeBrain hatching denotes provenance withholding rather than a structural gap; partial denominators are not rank-mixed. \textbf{b}, Source validation, target positive rate, hard- and probability-entropy, completed eligible cells, and single-class counts. Hatched N/A remains missing and is never replaced by zero.}
  \label{fig:benchmark-health}
\end{figure*}

The figure is a read-only view of the same frozen matrix and health ledger reported in Appendix~\ref{app:baseline-health}. It adds no baseline score, replacement denominator, or post hoc exclusion.

\FloatBarrier
\section{Sealed Coordinate-Perturbation Sensitivity}
\label{app:coordinate-perturbation}

This post-primary diagnostic leaves all trained parent checkpoints, prediction queues, and labels unchanged. Declared electrode coordinates were perturbed under two families: independent isotropic tangent-plane displacements and one global rigid head-frame rotation calibrated to the stated millimetre magnitude. The families were analyzed separately at 1, 2, 5, 10, and 20~mm. The final source table contains 50 target--condition units after averaging five registered seed repeats within each unit. Seventy-five tangent evaluations were reused read-only from a sealed legacy namespace; 175 evaluations and 35 MNE/fsaverage forward operators were added. The independent audit and module seal both returned PASS/COMPLETE with $P0=P1=0$.

\begin{figure*}[!t]
  \centering
  \includegraphics[width=0.98\textwidth]{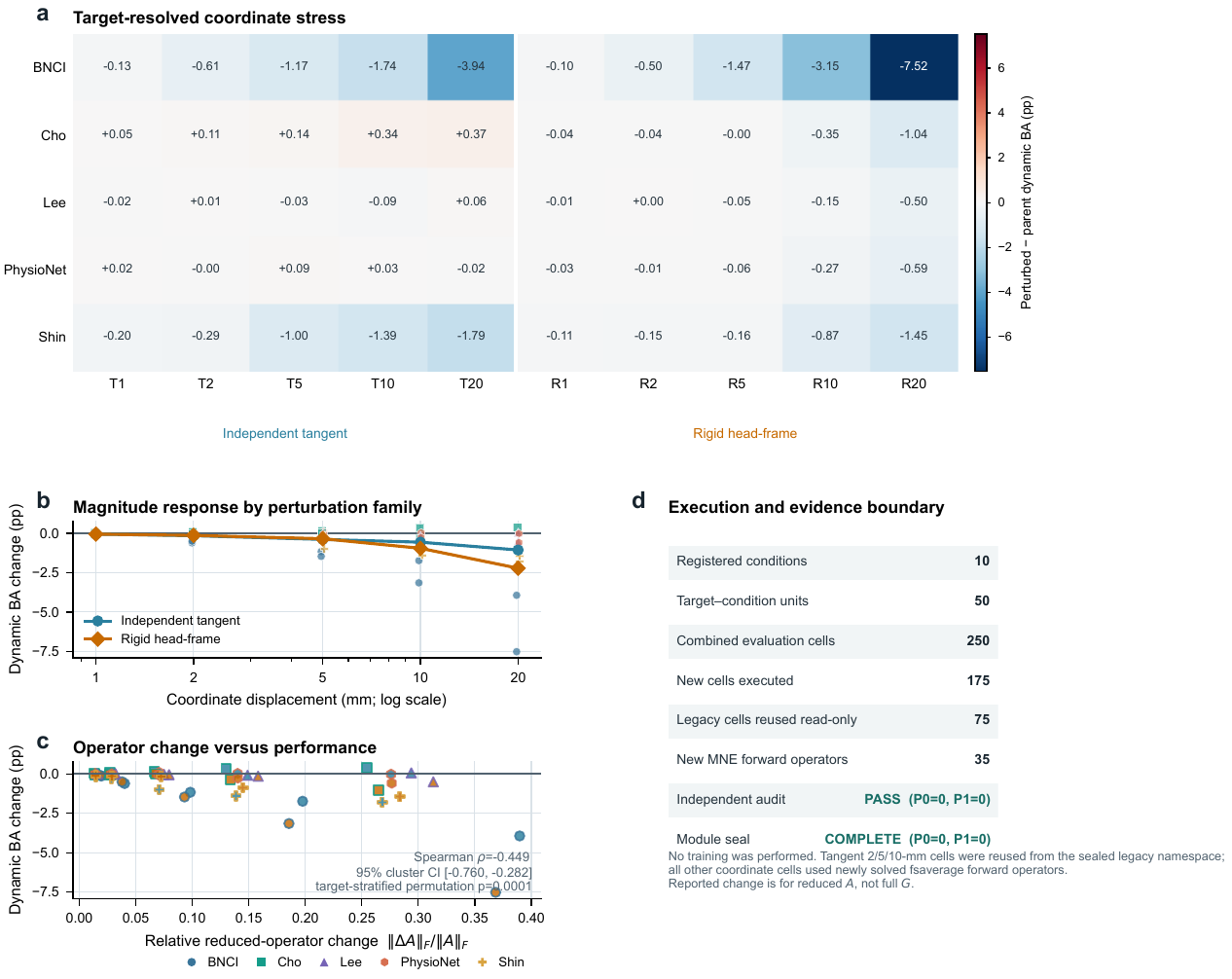}
  \caption{\textbf{Coordinate-perturbation sensitivity from sealed real-data evaluations.} \textbf{a}, Target-resolved dynamic BA change for independent tangent and rigid head-frame perturbations; T and R identify the families and the number gives millimetres. \textbf{b}, Magnitude profiles; points denote the five targets and lines equal-target means after averaging seeds. \textbf{c}, Reduced-operator relative Frobenius change versus dynamic BA change across 50 target--condition units (Spearman $\rho=-0.449$; target-cluster 95\% CI $-0.760$ to $-0.282$; target-stratified permutation $P<0.001$). \textbf{d}, Execution boundary. No training was performed, tangent and rigid conditions were not pooled, and the displayed operator change is for reduced $A$, not the full forward matrix $G$.}
  \label{fig:coordinate-perturbation}
\end{figure*}

\begin{table*}[h]
\centering
\caption{Equal-target coordinate-sensitivity summary. BA changes are percentage points from the frozen unperturbed parent; ranges are across the five held-out targets after averaging seeds. The operator column is the target mean of $\lVert\Delta A\rVert_F/\lVert A\rVert_F$.}
\label{tab:coordinate-perturbation}
\scriptsize
\setlength{\tabcolsep}{5pt}
\begin{tabular}{llrrrl}
\toprule
Family & Magnitude & Mean $\Delta$BA & Target range & Mean operator change & Artifact origin \\
\midrule
Independent tangent & 1 mm  & $-0.055$ & [$-0.195,+0.045$] & 0.0154 & new extension \\
Independent tangent & 2 mm  & $-0.157$ & [$-0.610,+0.107$] & 0.0309 & sealed read-only reuse \\
Independent tangent & 5 mm  & $-0.393$ & [$-1.165,+0.142$] & 0.0763 & sealed read-only reuse \\
Independent tangent & 10 mm & $-0.570$ & [$-1.736,+0.341$] & 0.1511 & sealed read-only reuse \\
Independent tangent & 20 mm & $-1.064$ & [$-3.935,+0.371$] & 0.2964 & new extension \\
Rigid head-frame & 1 mm  & $-0.058$ & [$-0.115,-0.009$] & 0.0155 & new extension \\
Rigid head-frame & 2 mm  & $-0.140$ & [$-0.502,+0.004$] & 0.0310 & new extension \\
Rigid head-frame & 5 mm  & $-0.348$ & [$-1.466,-0.004$] & 0.0765 & new extension \\
Rigid head-frame & 10 mm & $-0.957$ & [$-3.148,-0.146$] & 0.1526 & new extension \\
Rigid head-frame & 20 mm & $-2.220$ & [$-7.515,-0.504$] & 0.3014 & new extension \\
\bottomrule
\end{tabular}
\end{table*}

The monotone equal-target profiles and the negative rank association support sensitivity of this implemented reduced operator to coordinate error, but they do not imply a universal millimetre threshold. BNCI accounted for the largest losses, while several target--condition means remained near zero or slightly positive. This heterogeneity, the shared fsaverage anatomy, and the synthetic perturbation models limit interpretation to the five declared montages and the tested construction.

\section{Implementation and Registered Inference}
\label{app:implementation}

\paragraph{Active-set quality control.}
A channel is marked inactive if it contains a non-finite value, exceeds 5,000~$\mu$V in absolute amplitude, or has peak-to-peak amplitude below 0.1~$\mu$V. Trials are removed when more than 20\% of native channels are inactive or fewer than nine channels remain. The queue record binds the resulting mask to its subject and trial identifiers.

\paragraph{Template basis and source-only selection.}
For each hemisphere, the fsaverage forward model is restricted to the prespecified motor region. Gauge-projected lead fields from the five declared montages are stacked, and the four leading right-singular vectors are retained; each mode sign is set by its largest-magnitude loading. Source-library validation searches $\rho\in\{10^{-6},10^{-4},10^{-2},10^{-1}\}$ and $C\in\{0.01,0.1,1,10\}$, maximizing the unweighted mean subject-level BA. Ties favor the stronger ridge and then the smaller $C$. The dynamic branch uses AdamW (learning rate $10^{-3}$; weight decay $10^{-4}$), batch size 128, gradient clipping at 5, a maximum of 60 epochs, and patience 10. After selecting the epoch count, the branch is refitted on all source trials.

\paragraph{Exact same-parent operator ladder.}
The coordinate control evaluates the constant-free real spherical-harmonic basis $\{Y_{lm}:l\in\{1,2\},-l\leq m\leq l\}$ on unit electrode directions; degree zero is omitted because the active-reference quotient removes the common mode. The matched-spectrum control uses deterministic random left and right singular vectors with the physical operator's eight singular values. The row-shuffle control applies a no-fixed-point permutation to electrode rows, preserving the operator entries while breaking electrode--response correspondence. The wrong-anatomy control uses an equally sized visual-cortex leadfield solved with the same fsaverage head model. Each operator passes through the same active-reference projection and ridge inverse as FULL and is refitted on the same source libraries under inherited hyperparameters and epoch budgets. All 75 new fits were frozen before the 75 target evaluations began; physical, coordinate, and source-only sealed results were reused read-only. The independent composite audit covers 253 participants, 1,012 seed-averaged subject--control rows, and 5,060 subject--seed rows, and returned PASS ($P0=P1=0$). Source-only geometry remains a separately sealed sensitivity and is excluded from the four-control Holm family.

\paragraph{Calibration and uncertainty.}
Temperature calibration is fitted on source predictions only. Reliability curves pool trial counts into 15 fixed-width bins without smoothing; displayed bin intervals use 20,000 seed--target-cell bootstrap replicates. Scalar calibration metrics are computed per target and then averaged equally across targets. Main effects first average the five registered seeds within subject, sample the five target datasets with replacement, and then resample subjects within each selected target (20,000 replicates; seed 20260829). We report target-resolved effects, leave-one-target-out recomputations, and the worst target alongside the hierarchical interval. FI/SSI supplementary effects use 20,000 replicates with audit seed 20260824.

\paragraph{Synthetic known-source validation.}
Eight-region delayed Jansen--Rit-class networks \citep{jansen1995electroencephalogram} generate 3-s, 200-Hz trials after a 1.5-s burn-in. Five physiological parameters are sampled by stratified Latin hypercube for each target--seed pair. The registered channel budgets are 64, 32, 22, 16, and 8; SNR conditions are clean, +10, +5, 0, $-5$, and $-10$~dB. A geometry-conditioned Tikhonov inverse selects its ridge and orthogonal/scalar alignment on validation trials only. Test truth is used only for source correlation, NRMSE, spectral Wasserstein distance, one-step prediction NRMSE, and the registered failure gates. Shuffled-row and spectrum-matched operators are negative controls.

\begin{table*}[h]
\centering
\caption{Fixed implementation and selection settings. Any quantity marked ``source-selected'' is optimized without target signal, label, mask distribution, or empirical statistic.}
\label{tab:hyperparameters}
\small
\setlength{\tabcolsep}{4pt}
\begin{tabular}{lll}
\toprule
Component & Setting & Selection / role \\
\midrule
Preprocessing & notch; 0.5--45~Hz; 200~Hz & fixed for all datasets \\
Epoch & $[0.5,3.5)$~s; 600 samples & fixed \\
Hard channel gate & nonfinite; $>5000~\mu\mathrm{V}$; $<0.1~\mu\mathrm{V}$ p--p & fixed before inference \\
Trial gate & $\leq20\%$ inactive; at least 9 active & fixed before inference \\
Motor latent & 4 left + 4 right modes & fixed dimension $d=8$ \\
Transport ridge & $\rho\in\{10^{-6},10^{-4},10^{-2},10^{-1}\}$ & source-selected \\
Static logistic & $C\in\{0.01,0.1,1,10\}$ & source-selected \\
Static bands & 8--13, 13--30, 8--30~Hz & fixed; 24 features \\
SSM windows & 0.5~s; stride 0.25~s & fixed; 11 $\times$ 24 sequence \\
SSM state & 16 diagonal states & fixed architecture \\
Decay range & $(0.5,0.999)$ & bounded by parameterization \\
Residual range & $|r_{\rm SSM}|<2$ & bounded by parameterization \\
Optimizer & AdamW; lr $10^{-3}$; wd $10^{-4}$ & source fit only \\
Batch / clip & 128 / norm 5 & fixed \\
Epoch stopping & max 60; patience 10 & source-subject validation \\
Calibration & scalar temperature & source predictions only \\
Primary bootstrap & 20,000; seed 20260829 & dataset then subject \\
FI/SSI bootstrap & 20,000; seed 20260824 & target-stratified subject clusters \\
\bottomrule
\end{tabular}
\end{table*}

\FloatBarrier
\section{Efficiency Reporting Boundary}
\label{app:efficiency}

The efficiency record supports comparable GPU wall-clock summaries only for the executed neural fits; the classical pipelines did not record time over an equivalent scope. Table~\ref{tab:efficiency} therefore reports observed values and leaves unavailable timings unfilled. Static and dynamic time is not added because both share one fit record, and no universal speed ranking is inferred across different hardware or code paths.

\begin{table*}[h]
\centering
\caption{Available downstream parameter counts and source-fit times. Times are seconds per target--seed cell, reported as median [Q1,Q3]. Static and dynamic share one joint fit record and must not be summed. A dash marks unavailable comparable timing, not zero compute.}
\label{tab:efficiency}
\scriptsize
\setlength{\tabcolsep}{3pt}
\resizebox{\textwidth}{!}{%
\begin{tabular}{lrrrrl}
\toprule
Method & BA (\%) & Trainable & Fit time (s) & Timed cells & Scope / device \\
\midrule
\method dynamic & 60.865 & 1,281 & 289.93 [271.56, 413.96] & 25 & joint static--dynamic / CUDA \\
\method static & 59.806 & N/R & 289.93 [271.56, 413.96] & 25 & same joint record / CUDA \\
EEGConformer & 50.448 & 64,482 & 56.19 [35.06, 67.25] & 25 & selection + freeze / CUDA \\
EEGNet & 49.901 & 1,714 & 30.94 [24.57, 37.40] & 25 & selection + freeze / CUDA \\
CSP--LDA & 50.000 & 3 & --- & --- & comparable timer absent \\
FBCSP--LDA & 50.005 & 9 & --- & --- & comparable timer absent \\
Riemann--LR & 49.996 & 4 & --- & --- & comparable timer absent \\
\bottomrule
\end{tabular}
}
\end{table*}

Trainable-parameter counts follow the executed downstream protocol and exclude frozen checkpoint weights. A comparable static count was not recorded in the efficiency source, so Table~\ref{tab:efficiency} reports N/R rather than a post hoc reconstruction. VRAM is omitted because the runs did not share a common memory-measurement contract.

\FloatBarrier

\section{Access Contract and Audit Gates}
\label{app:audit}

Table~\ref{tab:audit-gates} separates source fitting, blind inference, one-time scoring, aggregation, and replay. Every transition requires complete, hash-matched outputs from the preceding stage. Completion further requires normal process exit, all expected cells, PASS decisions from the four group checks and aggregate replay, and $P0=P1=0$.

\begin{table*}[h]
\centering
\caption{Fail-closed execution and access separation. Each stage may read, but cannot rewrite, outputs accepted at an earlier gate.}
\label{tab:audit-gates}
\scriptsize
\setlength{\tabcolsep}{3pt}
\begin{tabularx}{\textwidth}{r>{\raggedright\arraybackslash}p{1.8cm}>{\raggedright\arraybackslash}p{4.3cm}>{\raggedright\arraybackslash}X>{\raggedright\arraybackslash}X}
\toprule
Step & Process & May read & Must emit & Failure rule \\
\midrule
1 & Source fit & source signals/labels; declared geometries & checkpoint + fit metadata & target store unavailable \\
2 & Fit barrier & all expected checkpoints & global fit manifest & any missing/hash drift stops run \\
3 & Blind inference & sealed fits; target signals; trial masks & probabilities/logits + queue IDs & labels unavailable \\
4 & Raw barrier & all expected raw predictions & raw-prediction manifest & any missing/hash drift stops score \\
5 & One-shot join & raw IDs; isolated target labels & paired score source & no fit artifact is writable \\
6 & Scorer & paired source; fixed metrics & cell metrics + subject metrics & no tuning or threshold selection \\
7 & Group audits & immutable fit/raw/score trees & independent group seals & any $P0/P1$ blocks aggregate \\
8 & Aggregate & passed score manifests only & target/equal-target summaries & structural N/A preserved \\
9 & Replay audit & aggregate sources in fresh process & independent aggregate seal & mismatch blocks completion \\
10 & Completion & all prior seals and process status & completion record & requires normal exit and $P0=P1=0$ \\
\bottomrule
\end{tabularx}
\end{table*}

\FloatBarrier
\section{Reproducibility and Source-Data Manifest}
\label{app:source-manifest}

The public source-data export contains 20 CSV files derived from 17 sealed source objects plus one hash manifest. Export does not fit models, rescore predictions, recalibrate probabilities, or change decisions. The manifest SHA-256 is \hashsplit{76c98cc6}{57c8d14d}{d27d013b}{b6783f41}{6edba1d9}{3d44b609}{87f3a4a1}{6e43bf4d}.

\begin{table*}[h]
\centering
\caption{Fourteen principal objects among the 17 immutable inputs behind the manuscript tables. The accompanying JSON manifest enumerates all 17 inputs and stores full digests, byte counts, paths, and aggregation rules; 12-character SHA-256 prefixes are shown here.}
\label{tab:source-data-manifest}
\scriptsize
\setlength{\tabcolsep}{4pt}
\begin{tabular}{llrl}
\toprule
Object & Scientific role & Rows / cells & SHA-256 prefix \\
\midrule
Figure2 subject pairs & Dynamic--BIOT held-out subjects & 253 rows & \texttt{0096b584532b} \\
Figure2 benchmark matrix & 18 methods $\times$ 5 targets & 90 rows & \texttt{c53c30d75146} \\
Figure2 target effects & two contrasts $\times$ 5 targets & 10 rows & \texttt{080901a117ed} \\
Hierarchical summary & 20,000-replicate primary inference & 1 JSON & \texttt{ad5334718806} \\
Strict target table & strict 15-method target matrix & 15 rows & \texttt{a46860b6bc08} \\
Secondary paired inference & 10 comparators $\times$ 3 metrics & 30 rows & \texttt{7a2d26ab3c01} \\
Locked cell metrics & registered target--seed reproducibility & 235 rows & \texttt{a101dd23b944} \\
FI/SSI target summary & 4 matched heads $\times$ 5 targets & 20 rows & \texttt{129e90ab64d6} \\
FI/SSI pooled inference & 4 heads $\times$ 4 metrics & 16 rows & \texttt{cc374de59310} \\
Few-shot target summary & 4 strategies $\times$ 3 fractions $\times$ 5 & 60 rows & \texttt{779746c6c249} \\
Known-source target ledger & 5 targets $\times$ 3 blocks & 15 rows & \texttt{a07b9141dc9d} \\
Known-source budget--SNR grid & 5 budgets $\times$ 6 SNRs & 30 rows & \texttt{c8777340972e} \\
Robustness family matrix & 8 executable + 2 structural N/A & 10 rows & \texttt{98d3ef25ec5f} \\
External registered summary & 2 absolutes + 2 contrasts & 4 rows & \texttt{79714d17cf28} \\
\bottomrule
\end{tabular}
\end{table*}

The exported tables do not recreate the scientific anchors. Their SHA-256 values remain \hashsplit{f08b2351}{f9e07de0}{c0078345}{4bce24b8}{a743153a}{9299556d}{a0da1a7c}{12534f38} for the five-target aggregate, \hashsplit{18f50901}{7357070e}{b2b71f30}{62a1c266}{210ec599}{85fa8dfb}{15ff730a}{792dceb6} for final statistics, and \hashsplit{c67d61b9}{842aa237}{c31873bd}{3130f846}{97537b2b}{51df68fa}{40f5532d}{995bb4f0} for the FI/SSI 60-score source.

\FloatBarrier

\section{Detailed Protocol-Fairness Ledger for Main and Matched Controls}
\label{app:fairness}

\begin{table*}[h]
\caption{Protocol-fairness ledger for access, adaptation, geometry, pretraining provenance, compute budget, and target denominator. Parameter counts include trainable downstream weights only; ``pre'' and ``post'' are defined relative to the model-fit lock.}
\centering
\scriptsize
\setlength{\tabcolsep}{2pt}
\begin{tabularx}{\textwidth}{>{\raggedright\arraybackslash}p{2.85cm}cccc>{\raggedright\arraybackslash}p{1.55cm}>{\raggedright\arraybackslash}Xrr}
\toprule
Method family & T/S/L & \shortstack{Tgt.\\tune} & \shortstack{Common\\ch.} & \shortstack{Tgt.\\geometry} & Pretraining & \shortstack{Executed\\source fit} & Trainable & Tgt. \\
\midrule
\rowcolor{GraftTeal!8}
\method static & none & no & no & pre, geometry only & none & 25 cells & N/R & 5/5 \\
\rowcolor{GraftBlue!8}
\method dynamic & none & no & no & pre, geometry only & none & 25 cells & 1,281 & 5/5 \\
\rowcolor{GraftOrange!10}
BIOT + LR & none & no & no & fixed 18-map & clean & 25 cells & 257 & 5/5 \\
CBraMod frozen + LR & none & no & no & checkpoint map & clean & 25 cells & 201 & 5/5 \\
CBraMod source-only full FT & none & no & no & checkpoint map & clean & 25 cells & full ckpt. & 5/5 \\
EEGPT + LR & none & no & no & fixed 19-map & overlap 1/5 & 25 cells & 513 & \partialden{4/5} \\
LaBraM + LR & none & no & no & fixed 19-map & overlap 1/5 & 25 cells & 201 & \partialden{4/5} \\
CodeBrain + LR & none & no & no & fixed 19-map & unresolved 5/5 & 25 cells & 201 & \provineligible{0/5} \\
Weighted CSP--LDA & none & no & yes & pre, channel names & none & 25 cells & 3 & 5/5 \\
Weighted FBCSP--LDA & none & no & yes & pre, channel names & none & 25 cells & 9 & 5/5 \\
Riemann tangent--LR & none & no & yes & pre, channel names & none & 25 cells & 4 & 5/5 \\
Compact EEGNet & none & no & yes & pre, channel names & none & 25 cells & 1,714 & 5/5 \\
Compact EEGConformer & none & no & yes & pre, channel names & none & 25 cells & 64,482 & 5/5 \\
ATCNet & none & no & yes & pre, channel names & none & 25 cells & N/R & 5/5 \\
ShallowFBCSPNet & none & no & yes & pre, channel names & none & 25 cells & N/R & 5/5 \\
Deep4Net & none & no & yes & pre, channel names & none & 25 cells & N/R & 5/5 \\
TFM & none & no & yes & pre, channel names & none & 25 cells & N/R & 5/5 \\
ODEBrain scratch & none & no & yes & pre, channel names & none & 25 cells & N/R & 5/5 \\
FI/SSI + tangent & none & no & no & post, exact-mask & none & 5 contrasts & head dep. & 5/5 \\
FI/SSI + EEGNet & none & no & no & post, exact-mask & none & 25 cells & head dep. & 5/5 \\
\bottomrule
\end{tabularx}
\end{table*}

Fit-time access excludes target-derived fitting, normalization, calibration, threshold selection, and early stopping for every method. ``T/S/L'' denotes target signal, empirical statistics, and labels; ``none'' means that none is used during fitting. Classical baselines operate on a predeclared common-channel intersection, whereas \method and FI/SSI retain each trial's native active set; foundation models use the fixed geometry maps required by their checkpoints. ``Effective targets'' is an eligibility denominator rather than an execution count: EEGPT, LaBraM, and CodeBrain each ran 25/25 cells, but known or unresolved pretraining overlap reduces their strict denominator. N/R denotes a comparable parameter count that was not recorded under the executed protocol and was not reconstructed after scoring.

\FloatBarrier

\section{Name, Code, and Result Identity}
\label{app:identity}

The manuscript name \method dynamic maps only to the code identifier \texttt{mosaic\_dynamic}; its static anchor is \texttt{mosaic\_static}. Table~\ref{tab:result-identity} prevents similarly named experimental namespaces from being treated as interchangeable. Full digests remain in the source-data manifest rather than the scientific table.

\begin{table*}[h]
\caption{Non-interchangeable result identities used in the manuscript.}
\label{tab:result-identity}
\centering
\small
\begin{tabularx}{\textwidth}{>{\raggedright\arraybackslash}p{3.0cm}>{\raggedright\arraybackslash}Xc}
\toprule
Paper role & Code identifier & Equal-target BA \\
\midrule
\rowcolor{GraftBlue!8}
Main dynamic method & \texttt{mosaic\_dynamic} & \textbf{60.865\%} \\
\rowcolor{GraftTeal!8}
Static anchor & \texttt{mosaic\_static} & \underline{59.806\%} \\
\rowcolor{GraftOrange!10}
Primary pretrained comparator & \texttt{biot} & \textbf{52.684\%} \\
\rowcolor{GraftPurple!10}
Exact same-parent layout-aware coordinate control & \texttt{mosaic\_dynamic\_exact\_}\linebreak\texttt{no\_physics\_missing\_v2} & 50.907\% \\
Matched-spectrum random operator & \texttt{matched\_spectrum\_random} & 51.167\% \\
Electrode-row-shuffled operator & \texttt{electrode\_row\_shuffle} & 49.624\% \\
Wrong-anatomy visual leadfield & \texttt{wrong\_anatomy\_visual} & 56.880\% \\
\bottomrule
\end{tabularx}
\end{table*}

The four operator controls are source-refitted children of \texttt{mosaic\_dynamic} under an identical parent contract. A later \texttt{graft\_parent\_FULL/A2--A8} namespace belongs to a different parent system and is not used for the paper's main claims or as a replacement for the 60.865\% result. Its identity record remains available in the reproducibility package; no A2--A8 effect is inferred here without a corresponding result table.

\FloatBarrier
\section{Fixed 17-Channel FI/SSI Target Montage}

The equal-information controls map each native active montage to
\begin{center}
\small
Fp1, Fp2, F7, F3, Fz, F4, F8, C3, Cz, C4, P3, Pz, P4, T7, T8, P7, P8.
\end{center}
Legacy labels T3/T4/T5/T6 are normalized to T7/T8/P7/P8. FI uses a minimum-norm field map with Tikhonov regularization $10^{-3}$; SSI uses spherical splines with regularization $10^{-7}$. The sequence is native active-set common-average reference, geometry map, fixed 17-channel common-average reference, fixed scaling by 100~$\mu$V, and clipping to $[-10,10]$ for EEGNet. Each exact active mask is mapped independently. The tangent reference point, covariance shrinkage, scaler, and classifier are fitted on source data only; no target-fitted recentering or calibration is used. Tangent targets use the registered single seed, whereas EEGNet targets use five seeds; aggregation preserves these original denominators.

\FloatBarrier
\section{Result Artifact Map}

The machine-readable artifact map records every locked object from which a quantitative figure was rendered, together with its retained denominator, source-data file, and output hash. It binds Figure~\ref{fig:architecture} to the Cho2017:0 replay; Figure~\ref{fig:main-results} to 253 paired participants, five targets, four exact same-parent controls, and the 20,000-replicate operator ledger; Figure~\ref{fig:external} to six selected real trials and all 60 external participants; Figure~\ref{fig:synthetic} to 925 registered known-source conditions; Figures~\ref{fig:real-eeg-atlas-1}--\ref{fig:real-eeg-atlas-3} to 15 label- and prediction-blind real trials; and Figure~\ref{fig:benchmark-health} to the strict target--method and baseline-health ledgers. The three-dimensional scaffold in Figure~\ref{fig:inverse-kernels} is marked visualization-only. The public map is limited to scientific artifacts, denominators, provenance records, and integrity hashes.

\end{document}